\documentclass[letterpaper, 10 pt, conference]{ieeeconf}  
\usepackage{color}
\usepackage{graphicx}
\usepackage{amsmath}
\usepackage{amssymb}
\usepackage{epsfig}
\usepackage{amssymb}
\usepackage{algorithm}
\usepackage{array}
\usepackage{mathtools}
\usepackage{color}
\usepackage[font=footnotesize, caption=false]{subfig}
\usepackage{url}
\usepackage{graphicx}
\usepackage{amsmath}
\usepackage{amssymb}
\usepackage{color}
\usepackage[noend]{algpseudocode}
\usepackage{overpic}

\usepackage[pagebackref=true,breaklinks=true,letterpaper=true,colorlinks,bookmarks=false]{hyperref}

\IEEEoverridecommandlockouts                              

\title{\LARGE \bf A robotic vision system to measure tree traits}

\author{Amy Tabb$^{1}$ and Henry Medeiros$^{2}$
\thanks{This paper's citation information is:  A. Tabb and H. Medeiros, ``A robotic vision system to measure tree traits," 2017 IEEE/RSJ International Conference on Intelligent Robots and Systems (IROS), Vancouver, BC, Canada, 2017, pp. 6005-6012. doi: \href{https://doi.org/10.1109/IROS.2017.8206497}{10.1109/IROS.2017.8206497} }
\thanks{Mention of trade names or commercial products in this publication is solely for the purpose of providing specific information and does not imply recommendation or endorsement by the U.S. Department of Agriculture.  USDA is an equal opportunity provider and employer.}
\thanks{A. Tabb acknowledges the support of US National Science Foundation grant number IOS-1339211.}
\thanks{We include an Erratum to this version of the paper. Please see Section \ref{erratum} for details.}
\thanks{$^{1}$USDA-ARS-AFRS Kearneysville, West Virginia, USA
        {\tt\small amy.tabb@ars.usda.gov}}%
\thanks{$^{2}$Marquette University, Electrical and Computer Engineering,
	Milwaukee, Wisconsin, USA
        {\tt\small henry.medeiros@marquette.edu}}%
}

\begin{document}

\maketitle
\thispagestyle{empty}
\pagestyle{empty}

\begin{abstract}

The autonomous measurement of tree traits, such as branching structure, branch diameters, branch lengths, and branch angles, is required for tasks such as robotic pruning of trees as well as structural phenotyping.  We propose a robotic vision system called the Robotic System for Tree Shape Estimation (RoTSE) to determine tree traits in field settings. The process is composed of the following stages: image acquisition with a mobile robot unit, segmentation, reconstruction, curve skeletonization, conversion to a graph representation, and then computation of traits.  Quantitative and qualitative results on apple trees are shown in terms of accuracy, computation time, and robustness. Compared to ground truth measurements, the RoTSE produced the following estimates: branch diameter (root mean-squared error $2.97$ mm), branch length (root mean-squared error $136.92$ mm), and branch angle (mean-squared error $31.07$ degrees).  The average run time was $8.47$ minutes when the voxel resolution was $3$ mm$^3$.

\end{abstract}

\section{INTRODUCTION}

This paper describes a system for autonomously sensing and describing the architecture of leafless trees in field settings.  This system is particularly useful for automation tasks related to managing fruit trees.  The most natural task for this system would be that of dormant pruning. In the dormant pruning task in the present day, workers selectively remove branches from fruit trees in order to optimize fruit production.  In some settings, they use pruners (sometimes called loppers) and ladders.  In others, they use pneumatically or battery-powered shears and ride on platforms.  However the configuration, the task is still manual, and requires humans to perform arduous work under harsh weather conditions since dormant pruning must be carried out during the winter months.  There has been significant interest in industry in the automation of dormant pruning.  In order to do so, it is necessary to autonomously detected the branching structure of trees, as pruning rules developed by horticulturists rely on specific measurements of the structure of the tree \cite{Lehnert2015automated}.

An alternative and equally challenging application for agricultural robotic vision systems is that of structural phenotyping.  An organism contains genetic information, called the genotype.  Its phenotype is the result of its genotype interacting with the environment.  In structural phenotyping, the structure of plants is sensed in order to build a map between genotype and phenotype, a field sometimes referred to as phenomics \cite{houle2010phenomics}.  While genotyping has become increasingly more automated, there has been little progress in the automation of phenotyping, particularly for structural traits acquired in outdoor scenarios. For tree crops, the state of the art consists of taking a sample of branches and manually measuring them with a flexible tape and a protractor, a labor-intensive and error-prone process. An automated system would allow greater numbers of trees to be measured in a more timely fashion, without the need for extra labor.

\begin{figure}[!ht]
\centering
	\includegraphics[width=\linewidth]{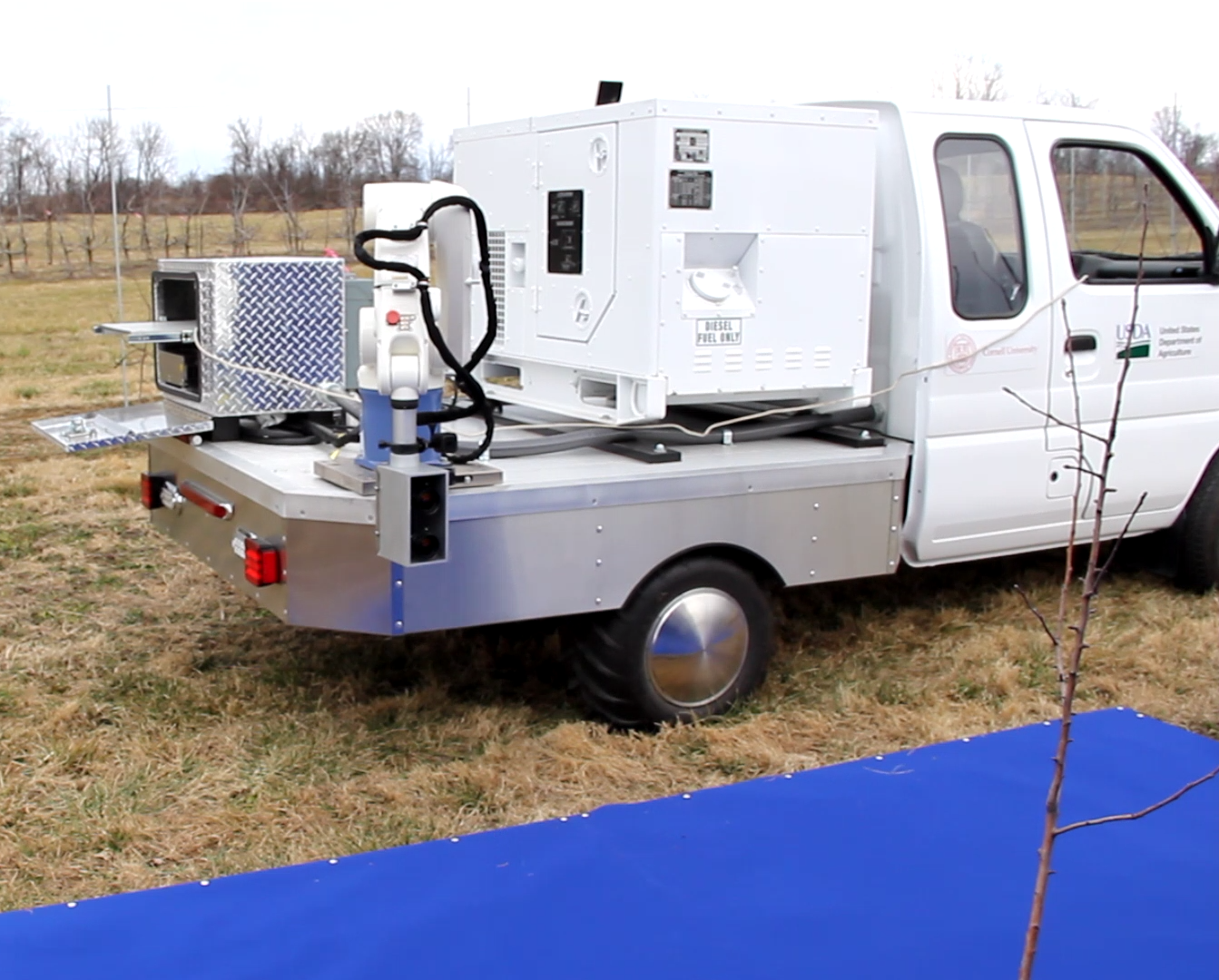}  
\caption{\textbf{[Best viewed in color.]} Robotic System for Tree Shape Estimation (RoTSE). The system consists of a Denso VS-6577G industrial arm and a generator mounted to a small truck.}
\label{fig:full_robot}
\end{figure}

\subsection{Contributions}

This paper describes a robotic system to determine tree structure autonomously (see Figure \ref{fig:full_robot}), which we denote as the Robotic System for Tree Shape Estimation (RoTSE) for the remainder of this paper. Its advantages over the state of the art are:
\begin{enumerate}
\item{This system is able to capture the following tree measurements with known accuracy: branch diameter (root mean-squared error $2.97$mm), branch length (root mean-squared error $31.07$mm), and branch angle (root mean-squared error $31.07$ degrees).}
\item{The system is able to determine tree shape and traits in $8.47$ minutes on average for small trees and does not require human intervention or hand-tuning.}
\end{enumerate}

\section{Related Work}

A variety of sensors and approaches have been used to sense tree structure.  For a detailed review relative to plant phenotyping, please see Li \textit{et al.} \cite{Li2014Review}. The sensors used are divided into the following categories: motion trackers, laser scanners, depth cameras, and color cameras.  

Motion trackers can be used to semi-automatically determine the locations of rigid, thick branches by manually positioning the sensors at the branch junction points of the tree \cite{sinoquet1997assessment,arikapudi2015orchard,vougioukas2016study}. Although this approach is accurate, it is labor-intensive and is restricted to measuring the positions and orientations of branches.

Tree measurement systems based on laser scanners \cite{medeiros2016modeling,livny2010automatic} can measure the trees in a more automated way, but their resolution is dependent on the distance to the branches. Hence, accurately and robustly estimating tree traits can become challenging, particularly for large point clouds. In addition, these systems are unable to utilize texture information from the trees to carry out reconstruction.

Structured-light depth cameras such as the Microsoft Kinect have been used to create models and estimate traits on landscape trees in \cite{liu2012image}, but they are limited to situations in which there is not direct sunlight exposure. More recent methods that rely on time-of-flight (ToF) RGBD sensors, have been showing promising results \cite{karkee2014identification,karkee2015method,elfiky2015automation,akbar2016novel,Chattopadhyay2016Measuring}. Although these methods overcome some limitations of lidar-based systems such as the inability to utilize texture, their resolution is also a function of distance and hence it can be difficult to apply them to larger structures. 

This work may be considered most close to the wine grape pruner of Botterill \textit{et al.} \cite{botterill2016robot}.  In \cite{botterill2016robot}, the vision component consists of a trinocular stereo color camera rig.  The three-dimensional structure of the vines and support structures is determined by segmenting images, estimating correspondences, and incrementally building a three-dimensional model of the row via bundle adjustment as the unit moves down the row.  While we also use color cameras and segment target regions in images, we capture many more than three camera positions from calibrated poses because of our use of a high-accuracy industrial robot (in this work, $56$ positions of two cameras are used), and as a result do not require bundle adjustment.  

\section{System description}

\begin{figure}[!ht]
	\centering
	\includegraphics[width=0.65\linewidth]{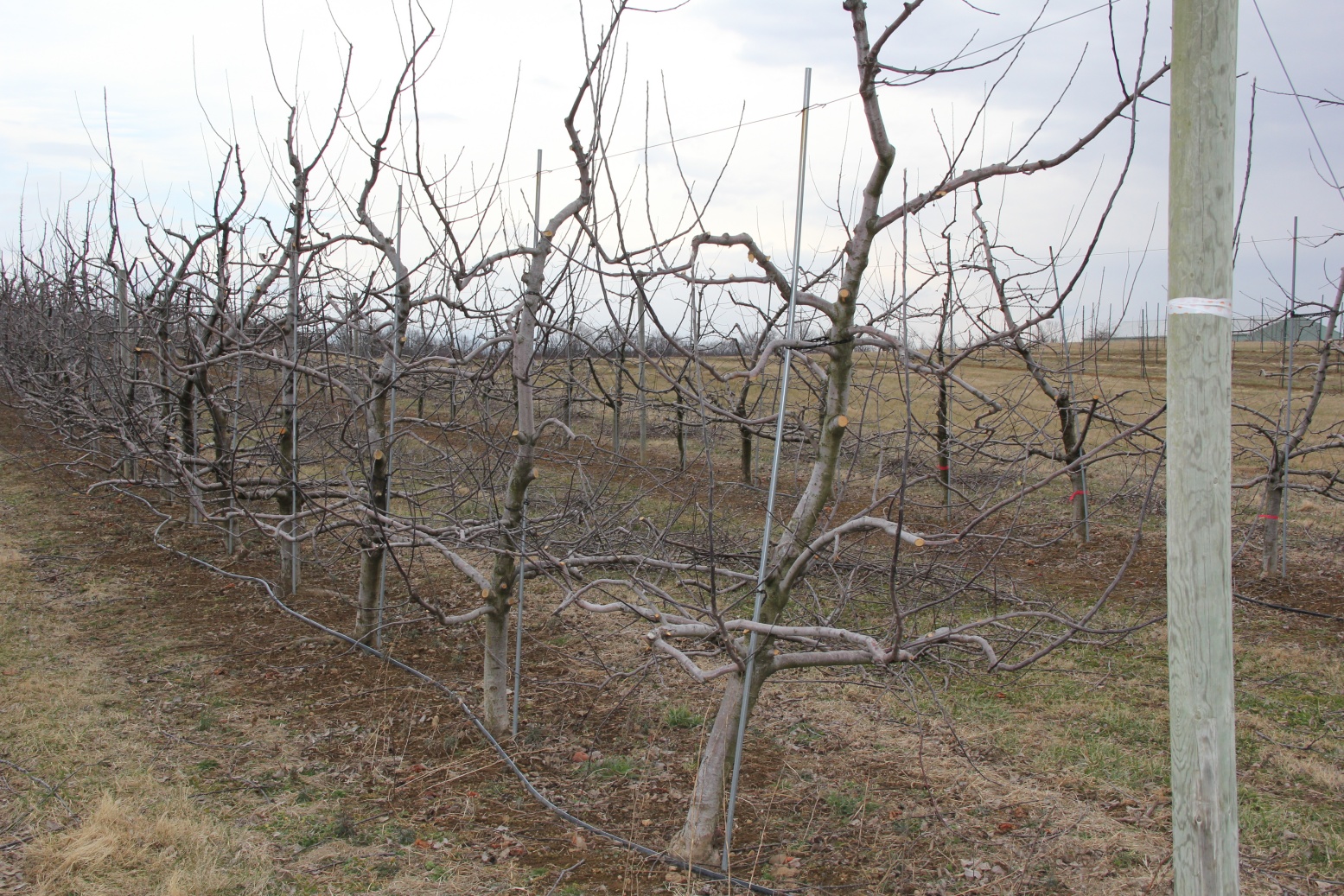}  
\caption{\textbf{[Best viewed in color.]} Example of a typical apple orchard. Trees are planted in rows, and supported with trellis.}
\label{fig:orchard}
\end{figure}

\begin{figure*}[!ht]
\centering
\begin{minipage}[b]{.15\linewidth}
\subfloat[]  
	    {  
	    \includegraphics[width=.95\linewidth]{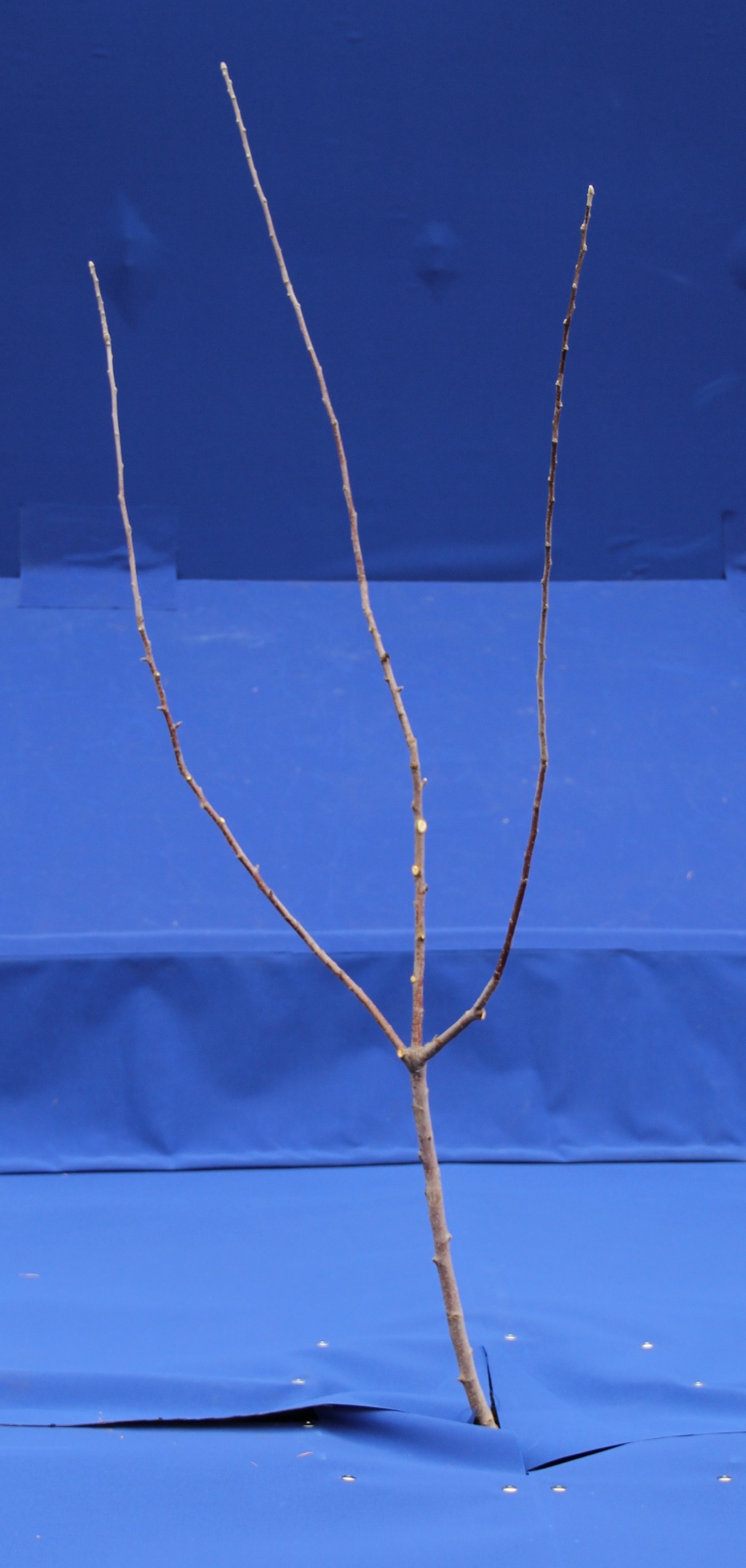}  

	    \label{sf:full_tree} } 
\end{minipage}
\begin{minipage}[b]{.20\linewidth}
    \subfloat[]  
	    {  
	    \includegraphics[width=\linewidth]{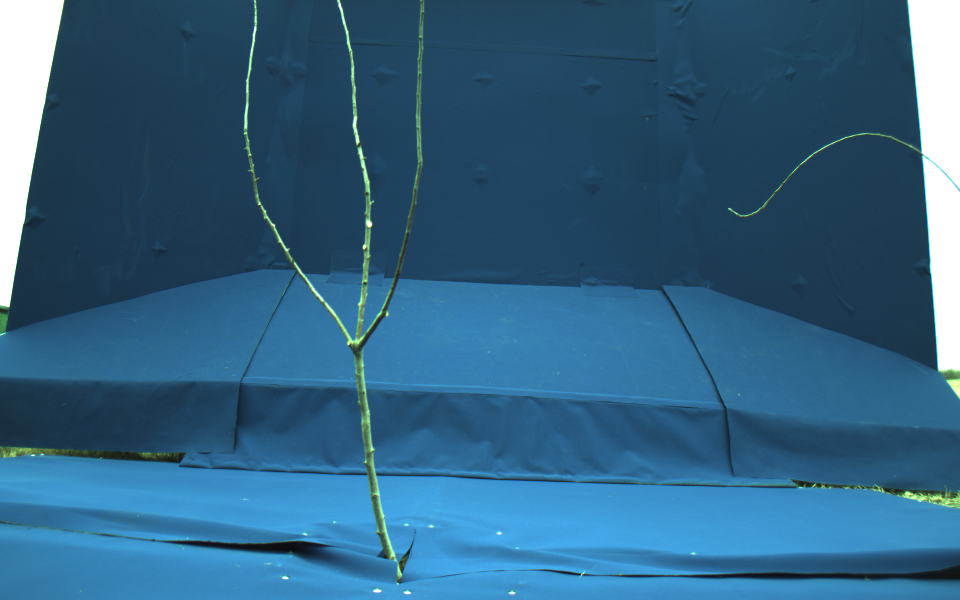}  

	    \label{sf:tree} } \\
    \subfloat[]  
	    {  
	    \includegraphics[width=\linewidth]{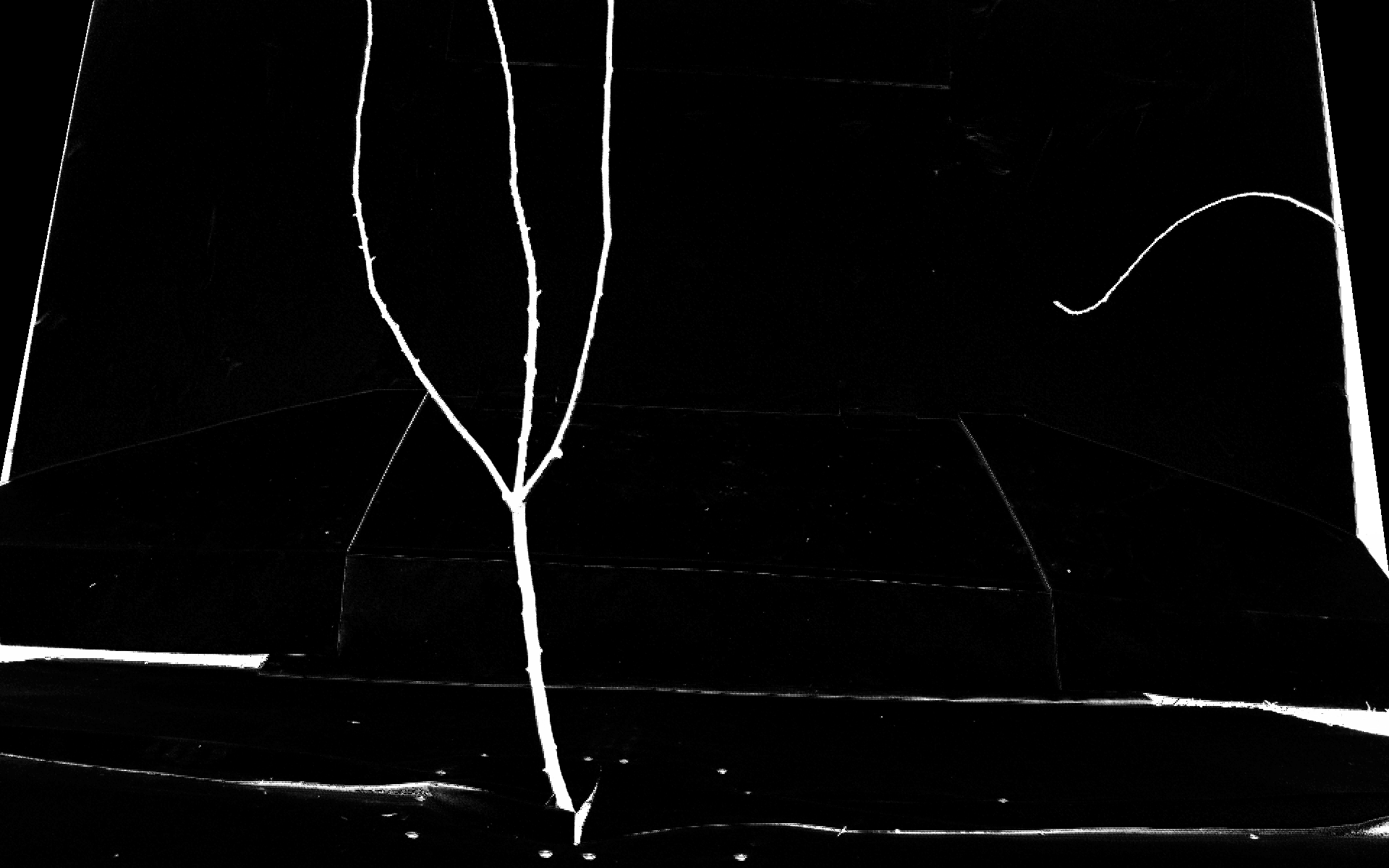} 
	    \label{sf:segmentation} }
\end{minipage}
\begin{minipage}[b]{.60\textwidth}
\subfloat[]  
	{  
	\includegraphics[width=0.28\linewidth]{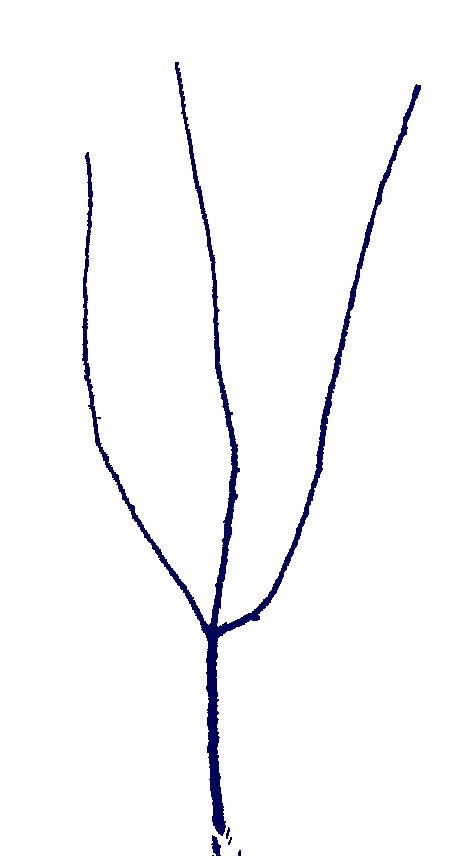}  

	\label{sf:recon} }
\subfloat[]  
	{  
	\includegraphics[width=0.28\linewidth]{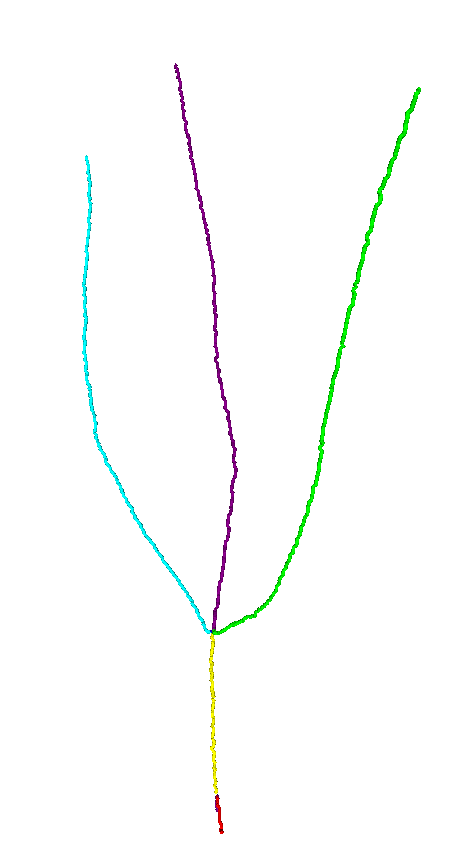}  

	\label{sf:skel} }
\subfloat[]  
	{ 
	\includegraphics[width=0.28\linewidth]{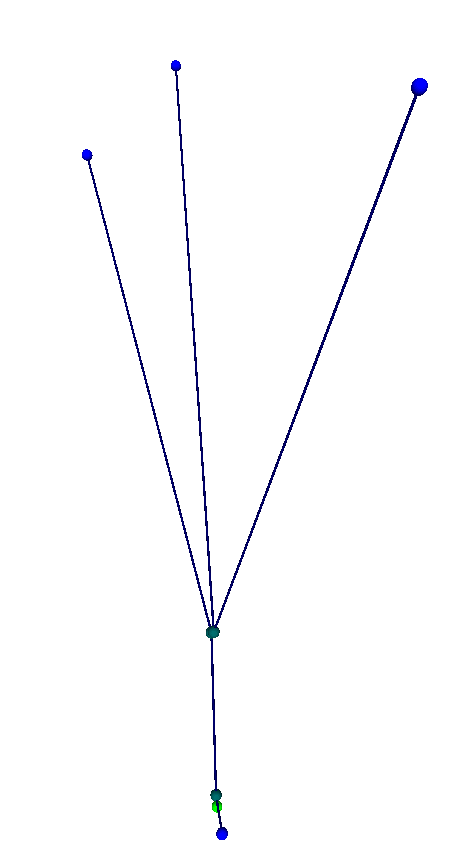}  
	\label{sf:graph_representation} }
\end{minipage}
\caption{\textbf{[Best viewed in color.]} \protect\subref{sf:full_tree} Full image of the tree acquired by a consumer camera.  This image is not used in the shape estimation process. \protect\subref{sf:tree} Image of the tree acquired by one of the cameras mounted on the robot. \protect\subref{sf:segmentation} Silhouette probability map of the image in \ref{sf:tree}.  \protect\subref{sf:recon}  Reconstruction of the tree volume using the silhouette probability maps.  \protect\subref{sf:skel} Curve skeleton computed from the reconstructed tree volume. \protect\subref{sf:graph_representation} Graph structure of the curve skeleton -- vertices are represented as spheres and edges as straight lines. Snapshots of the three-dimensional object files in this figure, and throughout this document were produced with the Meshlab viewer \cite{meshlab}.}
\label{fig:overview}
\end{figure*}

This section describes the hardware and software components of the system for tree shape estimation as well as their interdependencies and corresponding design choices. The system was designed based on the assumption that its robotic and vision components need to operate in an orchard setting, where trees are usually planted in rows and may or may not be supported by a trellis (see Figure \ref{fig:orchard}). The process for data acquisition is to park the robot unit such that the robot and cameras are facing the tree, and such that the background unit (Figure \ref{sf:background}) is behind the tree. Then the robot performs a series of movements within its workspace, acquiring images of the trees from a range of viewpoints.  When the movements are complete, both units are moved to the next tree. 

The images acquired at each stop of the mobile robot unit (MRU) are then provided as inputs to the software pipeline, which is responsible for computing all the tree traits observed in the image set. As explained in detail in Section \ref{ss:software}, the software pipeline consists of six completely autonomous steps. 

\subsection{Hardware components}
\label{ss:hardware}

\begin{figure}[!ht]
\centering
\subfloat[]  
	{  
	\includegraphics[height=1.4in]{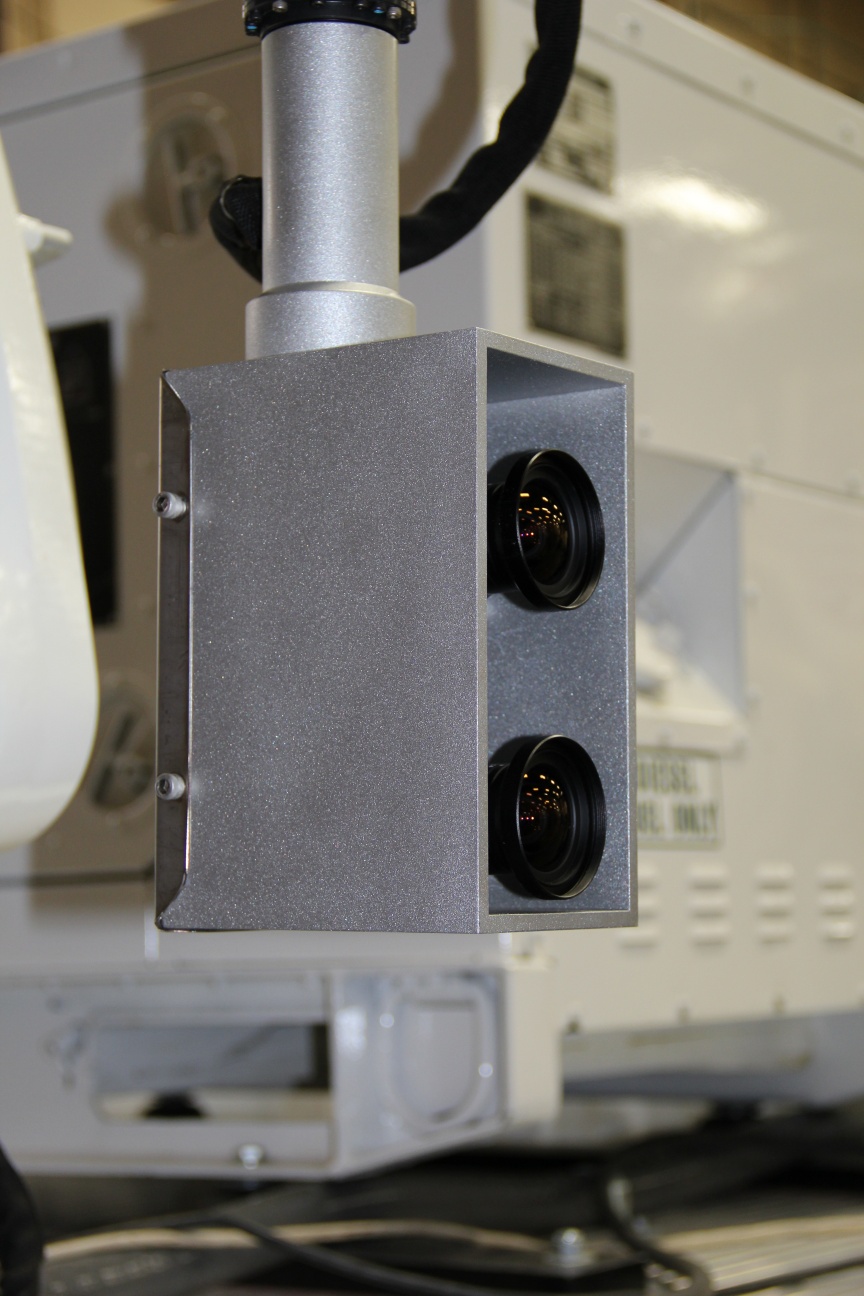}  
	\label{sf:cameras} }
\subfloat[]  
	{  
	\includegraphics[height=1.4in]{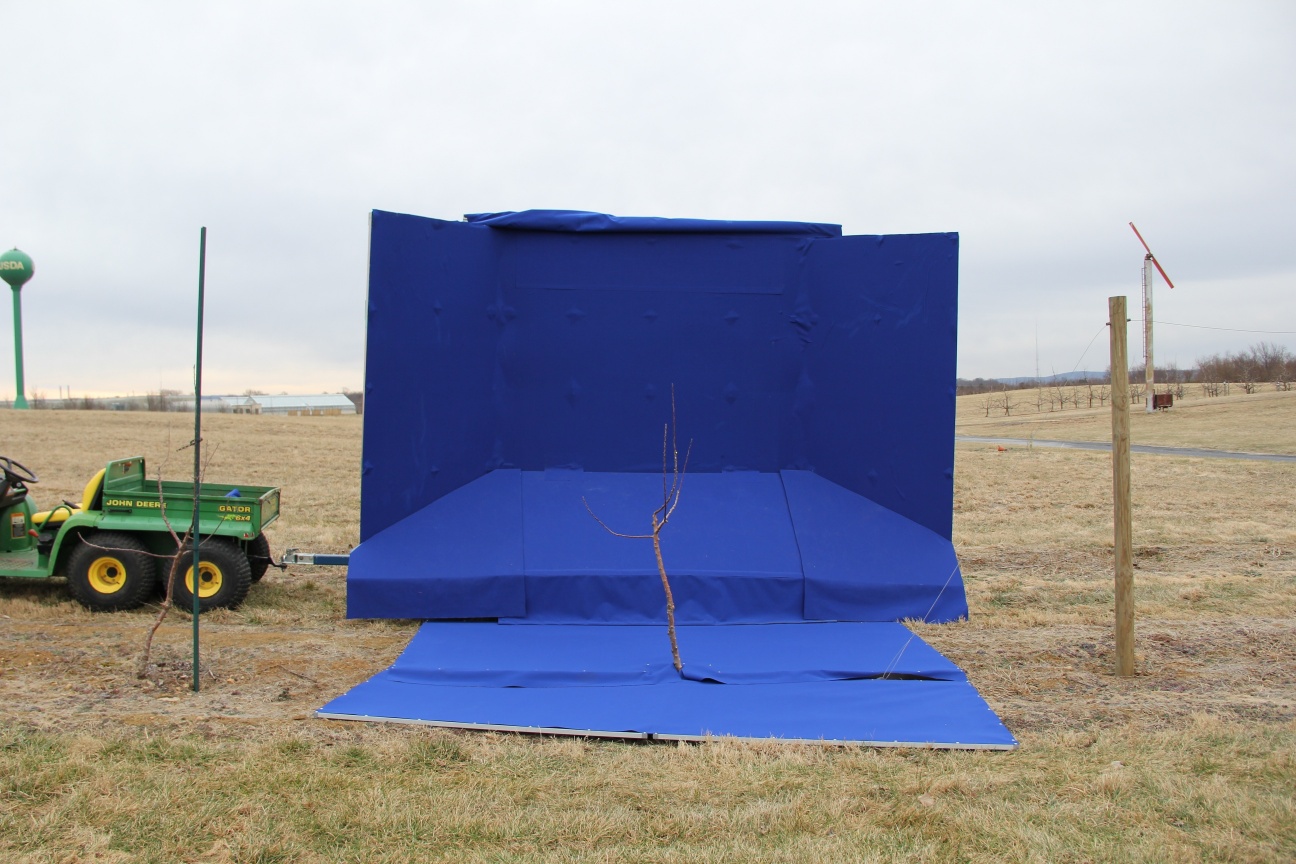} 
	\label{sf:background}  }
\caption{\textbf{[Best viewed in color.]} \protect\subref{sf:cameras} Close-up of the cameras and end-effector design.  \protect\subref{sf:background}  The background unit, which consists of a trailer for the vertical section, as well as a ground covering. }
\label{fig:hardware}
\end{figure}

The physical system consists of two units: the MRU, which carries the robotic arm with cameras mounted on the end-effector, and a background unit. The MRU consists of a small truck, with a robotic manipulator and a generator mounted on it, as shown in Figure \ref{fig:full_robot}.  The robot used is a Denso VS-6577G industrial arm with a 850mm reach.  The end-effector contains two color cameras (Figure \ref{sf:cameras}) with 8mm lenses; this choice was made so that multiple images could be captured at each stop of the robot, in order to minimize data acquisition time. The robot-camera calibration parameters are determined according to a modification of the calibration method for multiple cameras given in \cite{tabb2017solving} with the implementation available in \cite{tabb2017solving_dataset}.

The purpose of the background unit (Figure \ref{sf:background}) is to shield the cameras' view from trees that are not being modeled at the current time.  As previously mentioned, orchards are planted in rows, so when acquiring images of one leafless tree, other trees are visible.  The color of the background unit is blue, since that color is not naturally found in the orchard setting and hence facilitates the segmentation of the trees from the images.  

\subsection{Software components}
\label{ss:software}

As previously mentioned, at each stop of the MRU, the robot acquires a series of image pairs (i.e., one image per camera) at a number of predefined positions.\footnote{A video of this process is shown at \url{http://coviss.org/tabbmedeiros_rotse_iros17/}.} The sequence of images is fixed, and is determined empirically based on expected tree sizes and the available robot workspace and reach prior to experiments. The output after each stop of the MRU is a sequence of $n$ measurements, 
\begin{equation}
\mathbb{M}=\left\{ \left(I_{1}^{c},p_{1}\right), \left(I_{2}^{c},p_{2}\right), ..., \left(I_{n}^{c},p_{n}\right)\right\} _{c=1,2},\label{eq:measur}
\end{equation}
where $I_j^c$ is the $j$-th image acquired by camera $c$ and $p_j$ is the robot pose (i.e., position and orientation) at position $j$. Algorithm \ref{alg:overall} summarizes the processing steps carried out after each stop of the MRU. Figure \ref{fig:overview} shows an overview of the outputs of steps \ref{item:seg} to \ref{item:skeleton} of the algorithm. In the following sections, we explain each step in detail.

\begin{algorithm}
\caption{Computation of tree traits from measurements obtained by the MRU} \label{alg:overall}
\begin{algorithmic}[1]
\Require{Set of MRU measurements $\mathbb{M}$.}
\Ensure{Graph representation of tree traits.}
\State {Compute the external camera calibration parameters $\mathbf{R}_j^c$, $\mathbf{t}_j^c$ for each image $I_j^c$.}
\State {Segment tree versus non-tree regions according to the method described in \cite{tabb2017automatic}.} \label{item:seg}
\State {Reconstruct the tree volume using a shape-from-inconsistent-silhouette method detailed in \cite{tabb2013shape,tabb2014shape}.} \label{item:recon}
\State {Compute the curve skeleton from the noisy reconstruction in \ref{item:recon} according to the method in \cite{tabb2017fast}.}
\State {Convert the curve skeleton to a graph representation.}\label{item:skeleton}
\State {Compute the features of interest from skeleton representation.}\label{item:measure}
\end{algorithmic}
\end{algorithm} 

\subsubsection{Step 1: External camera calibration parameters computation}

The robot and two cameras are calibrated according to \cite{tabb2017solving}, which results in three $4 \times 4$  homogeneous transformation matrices (HTMs). The first transformation is the transformation from the robot base to the world coordinate system, denoted by $\mathbf{X}$. The second and third transformations concern transformations from the end-effector to the cameras.  The transformation from the end-effector to a camera $c$ is $\mathbf{Z}^c$, and since we have two cameras, $c=1,2$.\footnote{Note that these definitions are the inverse from much of the literature on robot-world, hand-eye calibration, and was done for specific results as outlined in \cite{tabb2017solving}.} 

Given a particular image $I_j^c$, we compute the HTM that corresponds to the external camera calibration parameters, meaning the rotation matrix and translation vector that transforms the world coordinate system to the camera's coordinate system for that stop of the robot arm.  We denote this matrix $\mathbf{A}_j^c$.  Let the transformation from the robot base to the robot hand when image $I_j^c$ was acquired be $\mathbf{B}_j^c$, then $\mathbf{A}_j^c$ is:
\begin{equation}
\mathbf{A}_j^c = \mathbf{Z}^c \mathbf{B}_j^c \mathbf{X}^{-1}
\end{equation}  
Internal camera calibration parameters, including those for distortion, are computed for each camera as a precursor to the robot-world, hand-eye calibration procedure.  In this manner, internal and external calibration information is available for all images.

\subsubsection{Step 2: Segmentation}

The method we use for reconstruction requires silhouettes or silhouette maps.  Although we use the background unit to control some aspects of the scene, illumination changes in the dynamic outdoor environment produces challenges when it comes to segmentation.  We use a method for segmentation that is described in \cite{tabb2017automatic}.  The method is a color-based one and locates probable blue background pixels and computes the optimal parameters for producing silhouette probability maps.  Silhouette probability maps are computed for each image $I_j^c$.

\subsubsection{Step 3: Reconstruction}

A method for shape-from-inconsistent-silhouette reconstruction is given in \cite{tabb2013shape}. In that work, it is assumed that each image has camera calibration information, and the reconstruction problem is formulated as a pseudo-Boolean optimization problem in a voxelized search area.  In \cite{tabb2014shape}, the run time of the method of \cite{tabb2013shape} was reduced by introducing a hierarchical version of the search; voxels start at a certain resolution, and the algorithm is run.  Then, voxels that were marked as occupied are included in the search area for the next round, as are neighbors within a certain distance, and these voxels are divided by an octtree.  This process continues according to the desired ending voxel size and number of octtree divisions, as set by the user. 

\subsubsection{Step 4: Skeletonization}

The resulting reconstruction from step 3 is characterized by a noisy surface, and depending on the ending voxel size, it can also be sparse.  The curve skeleton is considered as a locally one-dimensional curve that is roughly in the center of the object.  While there are many different approaches for computing the curve skeleton, we were not able to find any that could deal with the noisy nature and sparseness of our data.  Consequently, we compute the curve skeleton from the reconstruction in step 3 using our approach described in \cite{tabb2017fast}.  

\subsubsection{Step 5: Graph representation}
\label{sss:graph}

\begin{algorithm}
\caption{Conversion from curve skeleton to directed graph representation} \label{alg:skel_2_graph}
\begin{algorithmic}[1]
\Require{Set of skeleton segments $\mathbb{S}$.}
\Ensure{Directed graph $G = (V, E)$ and root $v_0$ representing the tree structure.}
\State $V = \lbrace \rbrace$, $E = \lbrace \rbrace$
\State $v_0$ = endpoint of $\mathbb{S}$ closest to the ground plane. \label{line:select_v_0}
\State $Q = \lbrace v_0 \rbrace$, where $Q$ is a queue.
\While {$|Q| > 0$ }
\State $v_a$ = first element of $Q$ and is removed from $Q$. 
\State $S$ = edges from  $\mathbb{S}$ that include $v_a$ as an endpoint.
\State $\mathbb{S} = \mathbb{S} - S$.
\For {each segment $s_j \in S$}
\State $s_j$ has two endpoints, $v_a$ and $v_b$. 
\State Create edge $e_j$: ($v_a$, $v_b$)
\State $v_b$ added to $Q$ if it is not already present.
\State $E = E \cup \lbrace e_j \rbrace$.
\EndFor
\EndWhile
\end{algorithmic}
\end{algorithm}

The output of the curve skeletonization step from step 4 is a set of undirected segments $\mathbb{S}$, each of which is made up of a set of a voxels. Each segment is indexed by its endpoints $v_a$ and $v_b$. This segment set is converted to a directed graph representation $G = (V, E)$ as described by Algorithm \ref{alg:skel_2_graph}. Initially, $V = \lbrace v_0 \rbrace$ and $E$ is empty.  The root of the graph, $v_0$ in line \ref{line:select_v_0} is selected by sorting the endpoints of the segments in $\mathbb{S}$ and choosing  the one that is closest to the ground plane in the world coordinate system. $v_0$ then becomes the first member of queue $Q$.  The graph is constructed from the segments by iteratively selecting the first item $v_a$ from $Q$, finding any segments in $\mathbb{S}$ incident to $v_a$, creating an edge $(v_a, v_b)$, and inserting it in $E$. The algorithm progresses until all vertices and edges have been explored. The result is a directed graph where $v_0$, in this application, represents the intersection of the tree with the ground.  Vertices with out-degree of zero represent terminal tips. 

\subsubsection{Step 6: Computation of features of interest}
\label{sss:features}

\begin{figure*}[!ht]
\centering
\subfloat[]  
	{ 
\begin{overpic}[width=0.23\textwidth]{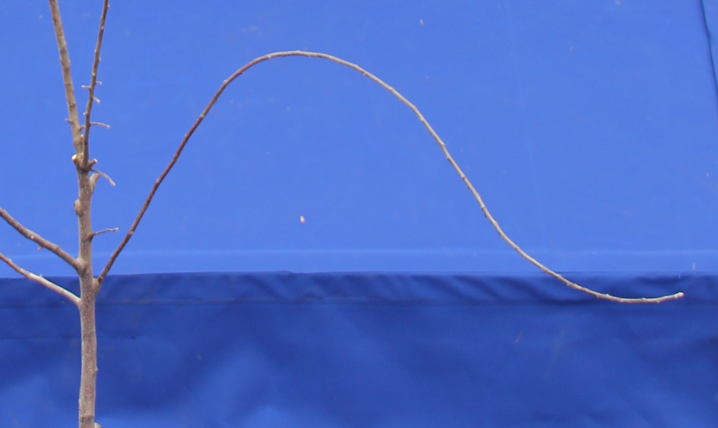}
\put(22.5,18){\color{yellow}\vector(-1,0){10}}  
\put(24,15){\small \color{yellow}Branch junction location}
\put(24,10){\small \color{yellow} $x, y, z$}
\end{overpic}  
}
\subfloat[]  
	{
\begin{overpic}[width=0.23\textwidth]{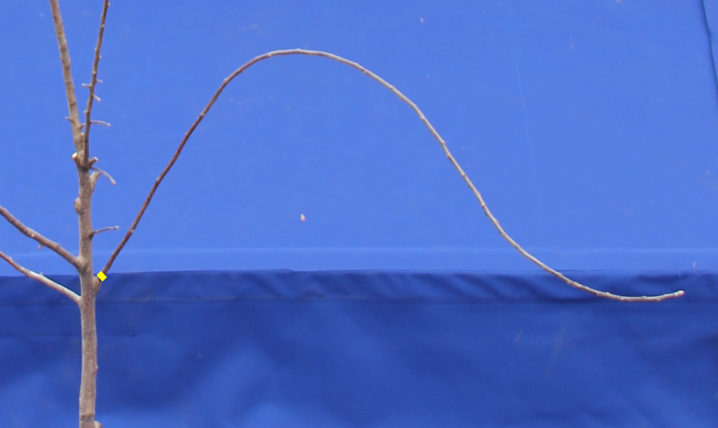}
\put(31.5,22){\color{yellow}\vector(-1,0){15}}  
\put(33,30){\small \color{yellow} Location}
\put(33,20){\small \color{yellow} of diameter}
\put(33,10){\small \color{yellow} measurement}
\end{overpic}  
\label{sf:branch_diameter}
}
\subfloat[]  
	{ 
\begin{overpic}[width=0.23\textwidth]{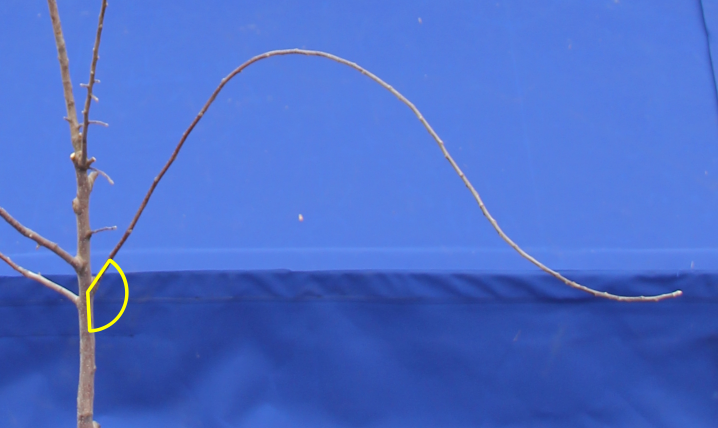}
\put(27,5){\color{yellow}\vector(-1,1){10}}   
\put(28,3){\small \color{yellow} Branch angle}
\end{overpic}  
}
\subfloat[]  
	{ 
\begin{overpic}[width=0.23\textwidth]{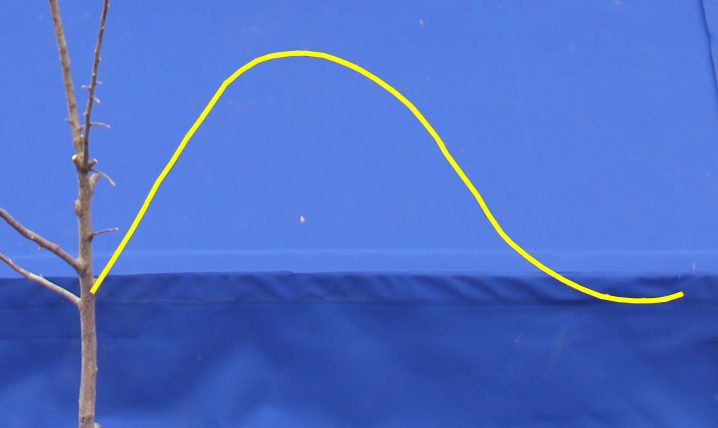}
\put(33,8){\color{yellow}\vector(-2,1){20}}  
\put(75,8){\color{yellow}\vector(2,1){20}}
\put(34,5){\small \color{yellow}Branch length}
\end{overpic}  
}
\caption{\textbf{[Best viewed in color.]} An illustration of the features computed by the RoTSE.}
\label{fig:features}
\end{figure*}
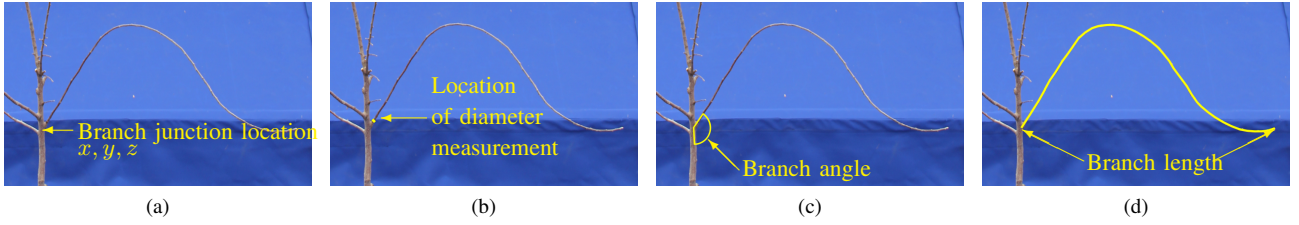

\providecommand{\norm}[1]{\lVert#1\rVert}

\begin{figure*}[!ht]
\centering
\subfloat[]  
	{ 
\begin{overpic}[width=0.24\linewidth]{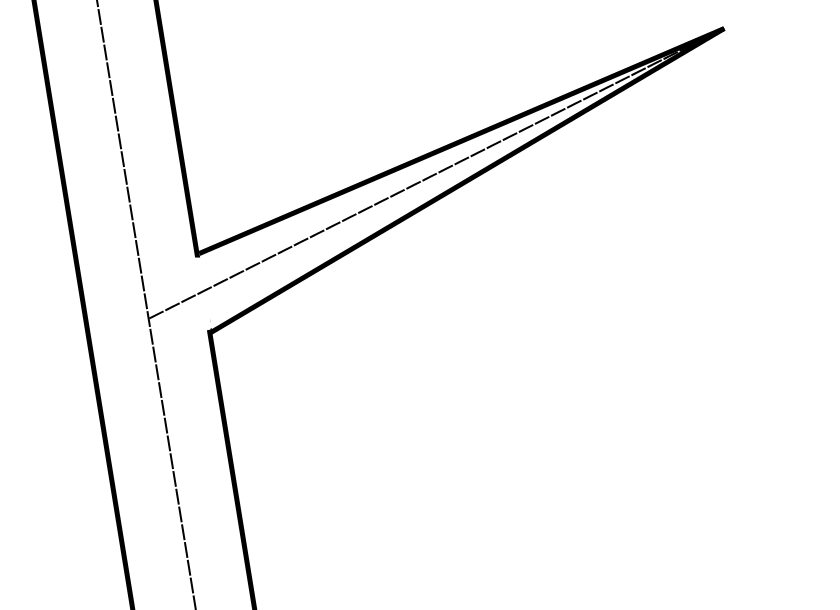}
\put(30,22){\color{black}\vector(-1,1){12}} 
\put(32,25){\small \color{black} Branch junction}
\put(32,17){\small \color{black} location}
\put(40,5){\color{black}\vector(-1,0){10}} 
\put(42,3){\small \color{black} Parent branch}
\put(25,60){\small \color{black} Child branch}
\put(35,60){\color{black}\vector(0,-1){12}} 
\end{overpic} 
\label{sf:branch_junction_drawing} 
}
\subfloat[]  
	{ 
\begin{overpic}[width=0.24\linewidth]{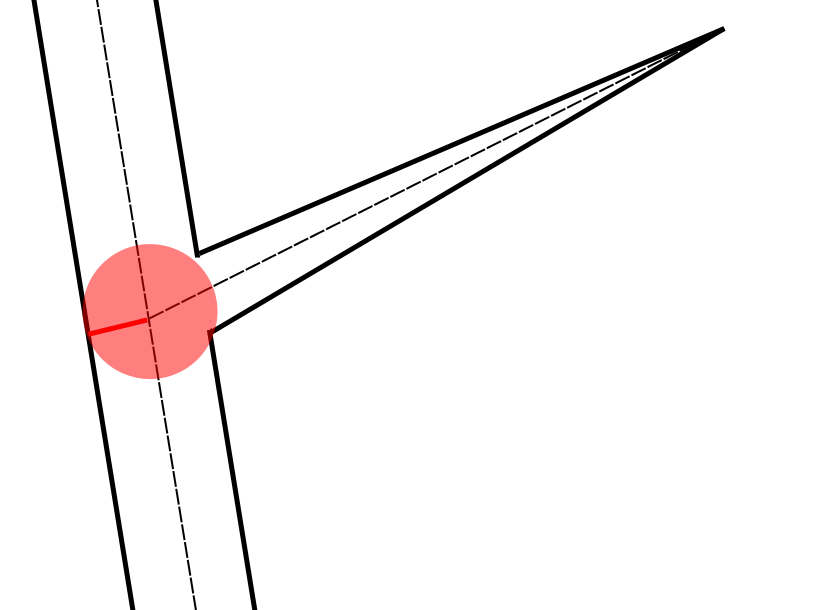}
\put(-1, 40){\small \color{black}$d_0$} 
\put(5,40){\color{black}\vector(2,-1){10}} 
\end{overpic}  
\label{sf:parent_r} 
}
\subfloat[]  
	{ 
\begin{overpic}[width=0.24\linewidth]{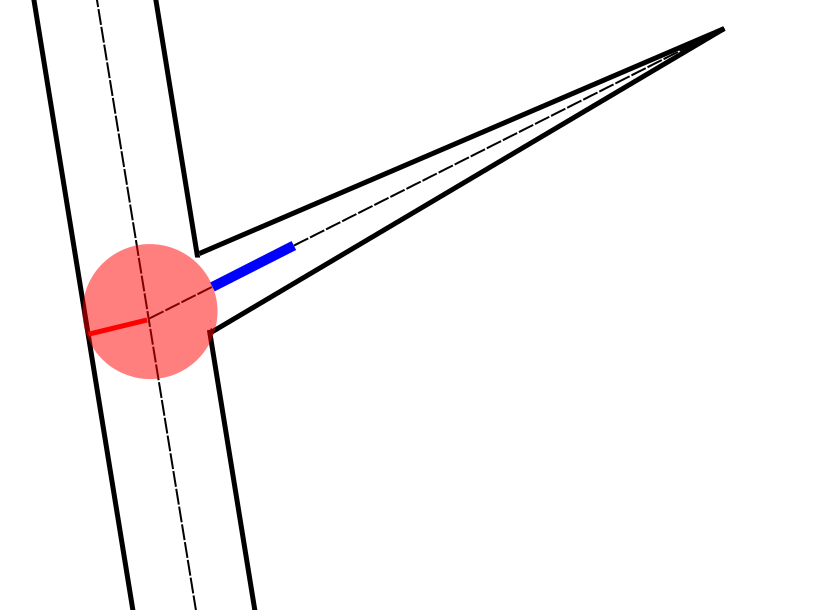}
\put(-1, 40){\small \color{black}$d_0$}
\put(5,40){\color{black}\vector(2,-1){10}} 
\put(40, 27){\small \color{black}$n_d$}
\put(38,30){\color{black}\vector(-1,1){10}} 
\end{overpic}  
\label{sf:child_n} 
}
\subfloat[]  
	{ 
\begin{overpic}[width=0.24\linewidth]{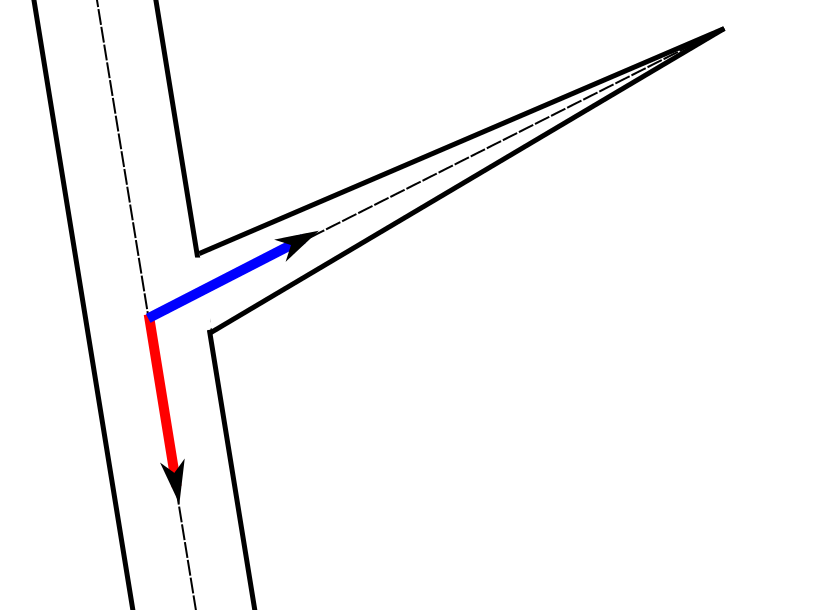}
\end{overpic} 
\label{sf:compute_angle}  
}

\caption{\textbf{[Best viewed in color.]} An illustration of how features are computed in Step 6. See text in Section \ref{sss:features} for a discussion.}
\label{fig:features_how}
\end{figure*}
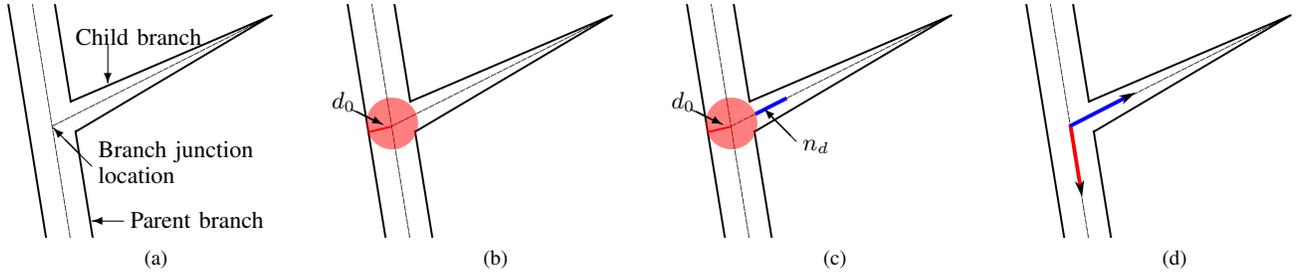

The outputs of the prior steps are used to compute the features of interest: branch junction locations, branch diameters, branch segment lengths, and branch angles.  For a visual illustration of the features see Figure \ref{fig:features}. Branch junction locations are determined from edge vertices with out-degree greater than zero. Vertices are centered on voxels, so determining the junction location consists of reading off the voxel's location within the voxel grid from the reconstruction step. The processes to compute the branch diameter, branch length, and branch angle are described with the help of Figure \ref{fig:features_how}. In this discussion, we assume that we are computing the features of the child branch, which is represented by the edge $e = (v_0, v_1)$, where $v_0$ corresponds to the branch junction location, which is determined as explained above, and $v_1$ is the other endpoint of the child branch. 

Once $v_0$ is known, the branch diameter is computed as follows. Branch junction location is the intersection of a branch with its parent branch inside the physical tree (Figure \ref{sf:branch_junction_drawing}), but we aim to find the branch diameter at a given distance from the branch junction location so that it reflects the horticultural practice of measuring branch diameters as in Figure \ref{sf:branch_diameter}. This process is shown in Algorithm \ref{alg:branch_diameter}. As part of the skeletonization step, distance labels $d_i$ are computed from every occupied voxel to the closest empty voxel using a linear time algorithm \cite{meijster2002general}.  In addition, the skeletonization step generates the ordered set $\mathbb{P}$, which corresponds to the curve segment vertices between $v_0$ and $v_1$ in order.  First, we let $d_0$ be the distance label representing the smallest distance to the surface of the reconstruction volume from $v_0$.  We make the assumption that the tree shape can be represented by a collection of spheres, and that $d_0$ is the radius of such a sphere centered at $v_0$. This traces out a sphere represented by the light red circle in Figure \ref{sf:parent_r}.  From there, we determine the voxel $\mathbb{P}_a$ on $\mathbb{P}$ that is $d_0$ distance away from $v_0$ (Lines \ref{line:start_while} - \ref{line:end_while}).  We then compute the average distance label of the next $n_d$ vertices to produce an average branch radius (Line \ref{line:n_d} and Figures \ref{sf:child_n} - \ref{sf:compute_angle}). $n_d$ is a parameter set by the user which has the purpose of removing locally occurring noise in the distance labels.  The diameter measurement $\hat{d}_e$ is given by twice the radius.

Branch angles are determined with another user parameter, $d_{angle}$.  It represents the distance from the branch junction where the branch angle should be measured, since branches are not typically straight in all regions.  Similarly to Lines \ref{line:start_while} - \ref{line:end_while} in Algorithm \ref{alg:branch_diameter}, the voxels that are $d_{angle}$ away on the paths from $v_0$ are located on the parent and child branches (represented by red and blue lines, respectively, in Figure \ref{sf:compute_angle}). Once these voxels are determined, vectors are generated and the angle between them is computed.

The branch length computation sums the Euclidean distances in between voxels on the path from $v_0$ to $v_1$, so that branch length $l$ is as follows:
\begin{equation}
 l = \sum_{i = 1}^{|\mathbb{P}| - 1} \norm{\mathbb{P}_i - \mathbb{P}_{i+1}}  
\end{equation}
where $\mathbb{P}_i$ is the $i^{th}$ vertex in $\mathbb{P}$.

\begin{algorithm}
\caption{Branch diameter determination} \label{alg:branch_diameter}
\begin{algorithmic}[1]
\Require{Path from $v_0$ to $v_1$ $\mathbb{P}$, distance labels $d_i$, parameter $n_d$}
\Ensure{Branch diameter estimate $\hat{d}_e$}
\State $d_0$ is the distance label of $v_0$
\State $a = 1$
\State $\mathbb{P}_a$ is the $a^{th}$ vertex in $\mathbb{P}$.
\While {$\norm{\mathbb{P}_a - v_0} < d_0$ } \label{line:start_while}
\State $a = a + 1$
\EndWhile \label{line:end_while}
\State $\hat{d}_e = 2\times \left(\frac{1}{n_d}\sum_{j = a}^{a + n_d} d_{\mathbb{P}_j}\right)$ \label{line:n_d}
\end{algorithmic}
\end{algorithm}

\section{Experiments}

We evaluate the performance of the RoTSE using data collected from twelve trees (denoted Trees A-L in subsequent sections) in the USDA-AFRS Orchard in Kearneysville, West Virginia. Performance is evaluated in terms of accuracy, computation time, and robustness. In all of our experiments, the trees were reconstructed using $56$ $1900 \times 1200$ pixel images. For the reconstruction step, the initial voxel size was set at $12$mm$^3$ and the final voxel size at $3$mm$^3$. The search region was $82$ million voxels, and it was the same for all of the trees. For the measurement of features step, $n_d = 5$ and $d_{angle} = 50$mm.  All of the results shown in this paper were generated on a workstation with one 12-core processor and 192 GB of RAM.  The following subsections describe the experimental procedures in detail.  

\subsection{Accuracy}

Accuracy is evaluated by comparing the tree traits computed by our system with manual measurements obtained from trees in the orchard. Since ground truth generation is labor-intensive, we limit our manual measurements to two trees (Trees A and G). Our evaluation focuses on the following parameters (see Figure \ref{fig:features}): number of branches, branch diameter, branch length, and branch angle. Branch location is difficult to measure in practice without resorting to equipment such as used in \cite{sinoquet1997assessment,arikapudi2015orchard,vougioukas2016study}. Hence, we omit that specific parameter from our evaluation.

\begin{figure}[!ht]
\centering
\subfloat[Tree A]  
	{ 
\begin{overpic}[trim = 3.5cm 5cm 6.5cm 5cm,clip=true,width=0.48\linewidth]{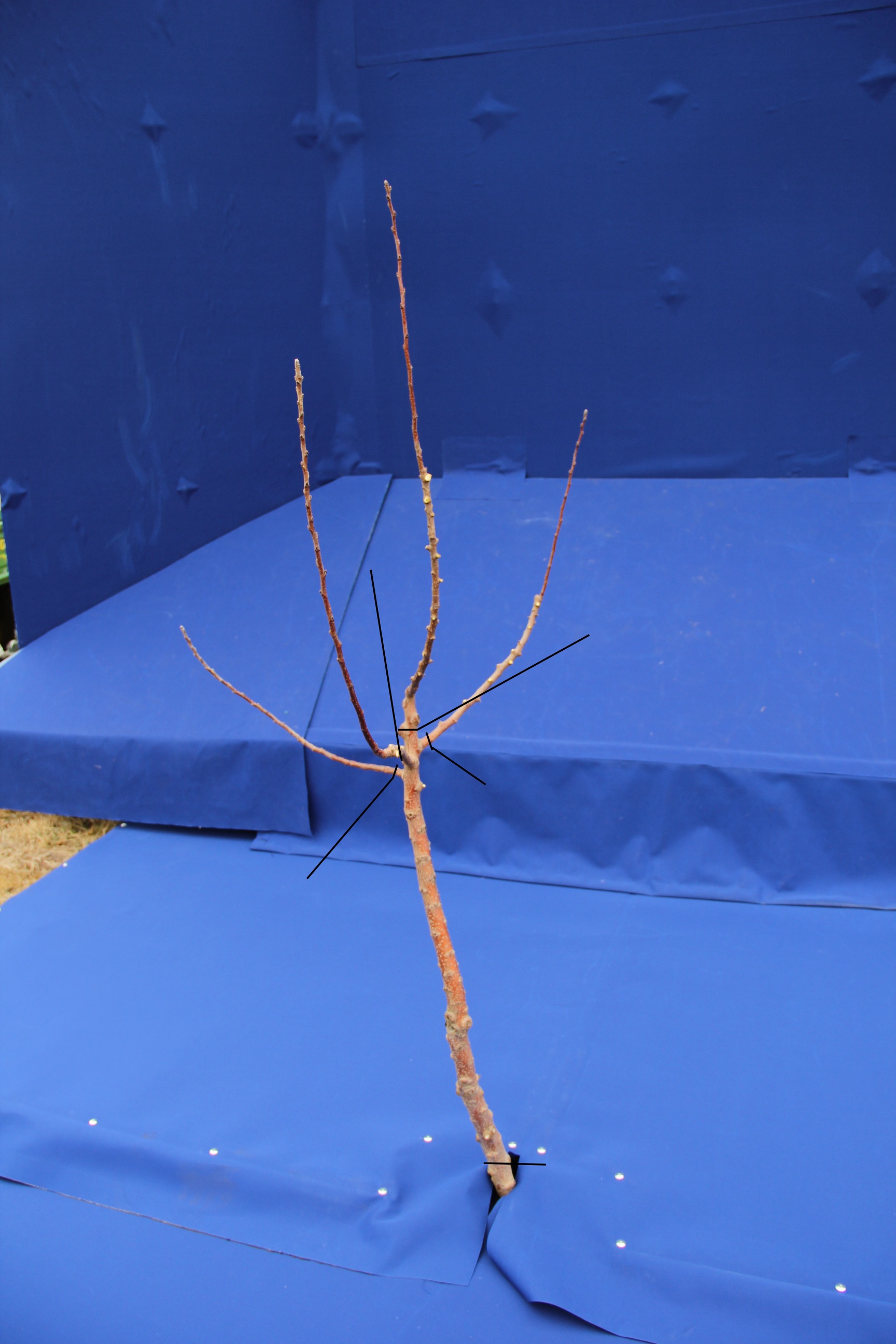}  
\put(15, 15){\small \color{white}Branch 0}
\put(3, 55){\small \color{white}Branch 1}
\put(10, 77){\small \color{white}Branch 2}
\put(20, 92){\small \color{white}Branch 3}
\put(40, 75){\small \color{white}Branch 4}
\put(42, 8){\small \color{white}34.5mm}
\put(15, 30){\small \color{white}14.1mm}
\put(22, 60){\small \color{white}17.5mm}
\put(35, 40){\small \color{white}16.1mm}
\put(45, 52){\small \color{white}23.0mm}
\end{overpic}  
	} 
\subfloat[Tree G]  
	{  
\begin{overpic}[trim = 7.5cm 10cm 2.5cm 0cm,clip=true, width=0.48\linewidth]{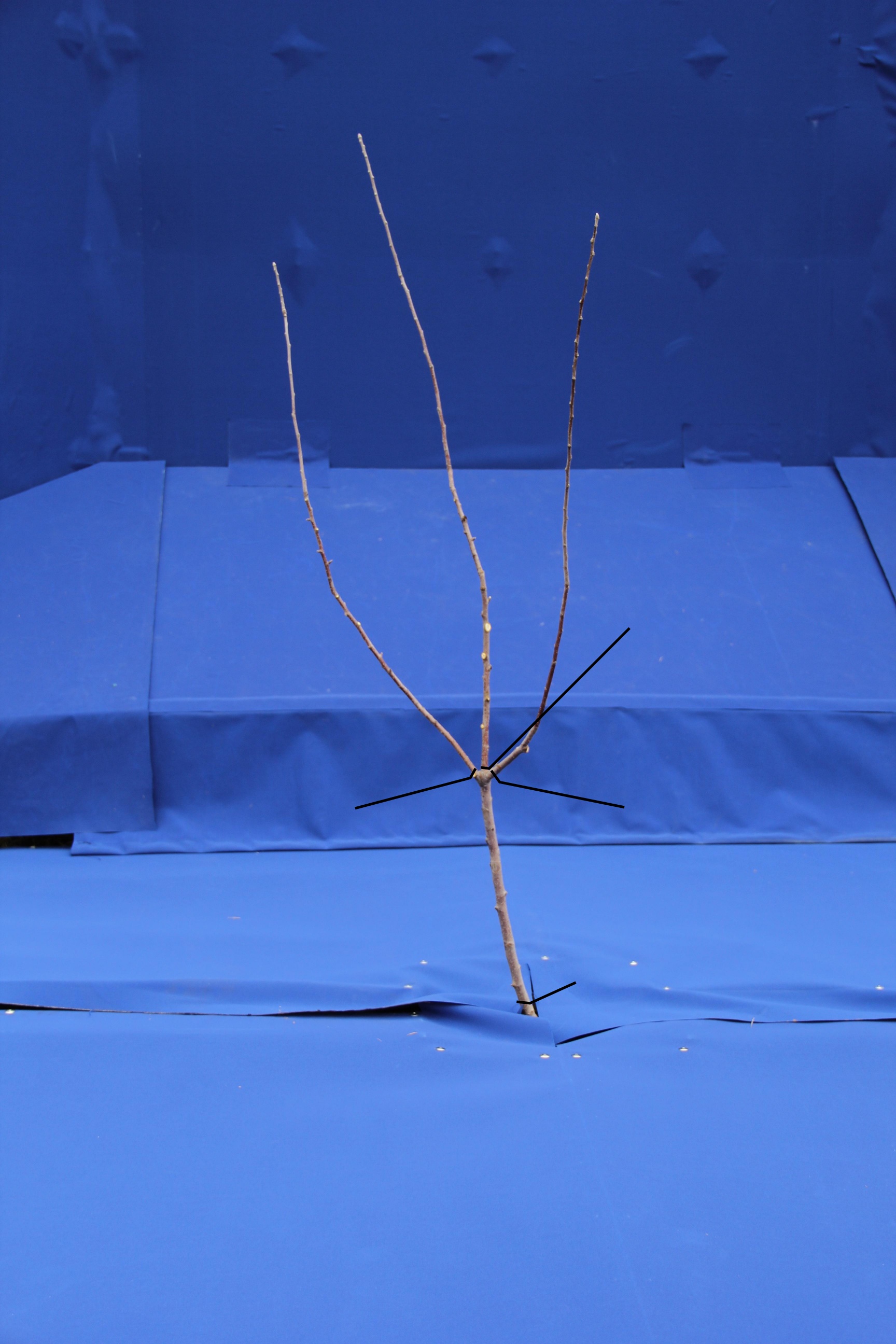} 
\put(15, 20){\small \color{white}Branch 0}
\put(0, 70){\small \color{white}Branch 1}
\put(12, 90){\small \color{white}Branch 2}
\put(37, 85){\small \color{white}Branch 3} 
\put(39, 17){\small \color{white}18.9mm}
\put(7, 32){\small \color{white}8.7mm}
\put(43, 48){\small \color{white}10.4mm}
\put(42, 32){\small \color{white}12.7mm}
\end{overpic} 
	}
\caption{\textbf{[Best viewed in color.]} Trees used in the accuracy evaluation along with branch identifiers and their corresponding diameters.}
\label{fig:tree_gt}
\end{figure}

\begin{figure}[!h]
\centering
	\includegraphics[width=\linewidth]{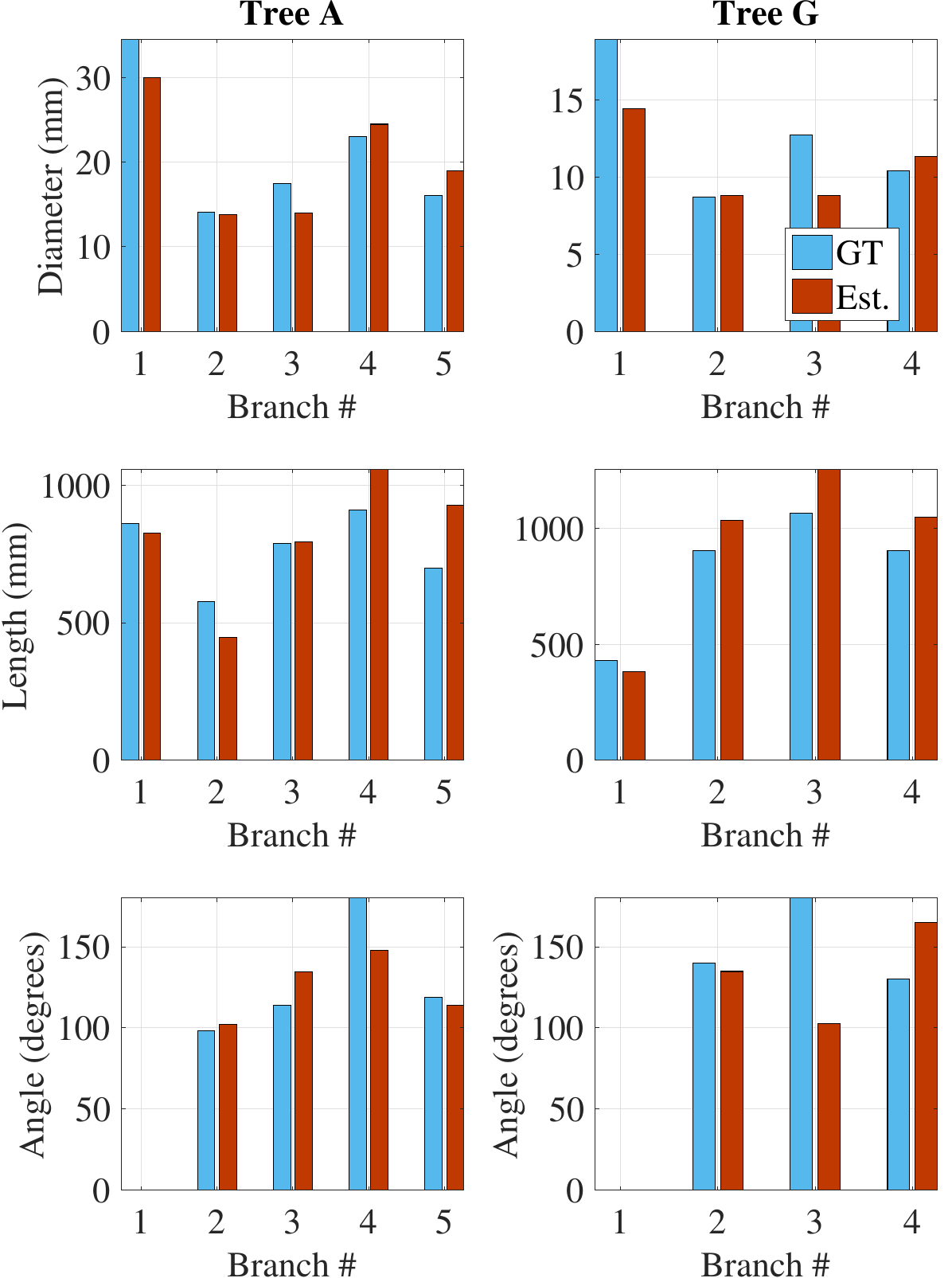}  
\caption{Quantitative accuracy results. GT is the ground truth and Est. are the measurements estimated by our system.}
\label{fig:accuracy_res}
\end{figure}

Figure \ref{fig:tree_gt} shows the two trees used in the accuracy evaluation. The trees for this quantitative comparison were simplified so that they consisted of only a trunk and primary branches; higher-order branches were removed. The labels in the figures correspond to the branch numbers and their corresponding diameters (other parameters are omitted from the figure for clarity). When the manual measurements were acquired, each branch was given a label and the measurements were noted on a photograph of the tree.  Visual inspection with a 3D model viewer was used to determine the mapping from the RoTSE labeling and the manually labeled branches so that measurement error could be reported. Figure \ref{fig:accuracy_res} shows the accuracy results for the trees shown in Figure \ref{fig:tree_gt}. Table \ref{table:accuracy} summarizes the error statistics.

\begin{table}[!h]
\caption{Summary of the accuracy of RoTSE for 2 trees. RMSE is the root mean squared error and STD is the standard deviation of the error.}
\begin{center}
{
\begin{tabular}{l||c|c|c}
\hline 
 & Diameter (mm) & Length (mm) & Angle (degrees)\tabularnewline
\hline 
\hline 
RMSE & 2.97 & 136.92 & 31.07\tabularnewline
\hline 
STD & 2.85 & 124.51 & 32.18\tabularnewline
\hline 
\end{tabular}
}
\end{center}
\label{table:accuracy}
\end{table}

\subsection{Computation time}

The computation time for each step of the proposed approach is shown in Table \ref{table:computation_time} for all the twelve trees. The average time to compute the tree features is approximately $8.5$ minutes in addition to approximately $1$ minute for data acquisition and the time to move the platform to the next tree. Note that feature computation for a given tree can be performed in parallel with data acquisition for the following tree so that the overall time can be amortized as multiple trees are measured.

\begin{table}[t]
\caption{Computation time for the RoTSE, 12 trees}
\begin{center}
\resizebox{\linewidth}{!}
{
\begin{tabular}{  l | l | l | p{20mm} | p{15mm} }
\hline
	Tree ID & Segmentation (s) & Reconstruction (s) & Skeletonization, graph rep., feature measurement (s) & Total time (minutes) \\ \hline
	A & 15.59 & 333.72 & 4.17 & 5.89 \\ \hline
	B & 19.78 & 372.14 & 3.73 & 6.59 \\ \hline
	C & 17.30 & 395.68 & 3.84 & 6.95 \\ \hline
	D & 14.88 & 409.14 & 5.32 & 7.16 \\ \hline
	E & 20.43 & 445.29 & 3.74 & 7.82 \\ \hline
	F & 16.02 & 476.22 & 4.54 & 8.28 \\ \hline
	G & 22.17 & 471.04 & 3.82 & 8.28 \\ \hline
	H & 16.10 & 517.11 & 3.92 & 8.95 \\ \hline
	I & 19.35 & 523.08 & 4.75 & 9.12 \\ \hline
	J & 17.14 & 527.36 & 4.02 & 9.14 \\ \hline
	K & 19.69 & 667.07 & 5.38 & 11.54 \\ \hline
	L & 20.31 & 690.50 & 4.33 & 11.92 \\ \hline
    \hline
	{\bf Average} & {\bf 18.23} & {\bf 485.70} & {\bf 4.30} & {\bf 8.47} \\ \hline
	{\bf Stdev} & {\bf 2.34} & {\bf 109.07} & {\bf 0.59} & {\bf 1.84} \\ \hline
\end{tabular}
}
\end{center}
\label{table:computation_time}
\end{table} 

\subsection{Robustness}

\begin{figure*}[!ht]
\centering
\begin{minipage}[b]{.24\linewidth}
\subfloat[Full image of tree B]  
	    {  
	    \includegraphics[trim = 5cm 15cm 10cm 7.5cm,clip=true, height=0.28\textheight]{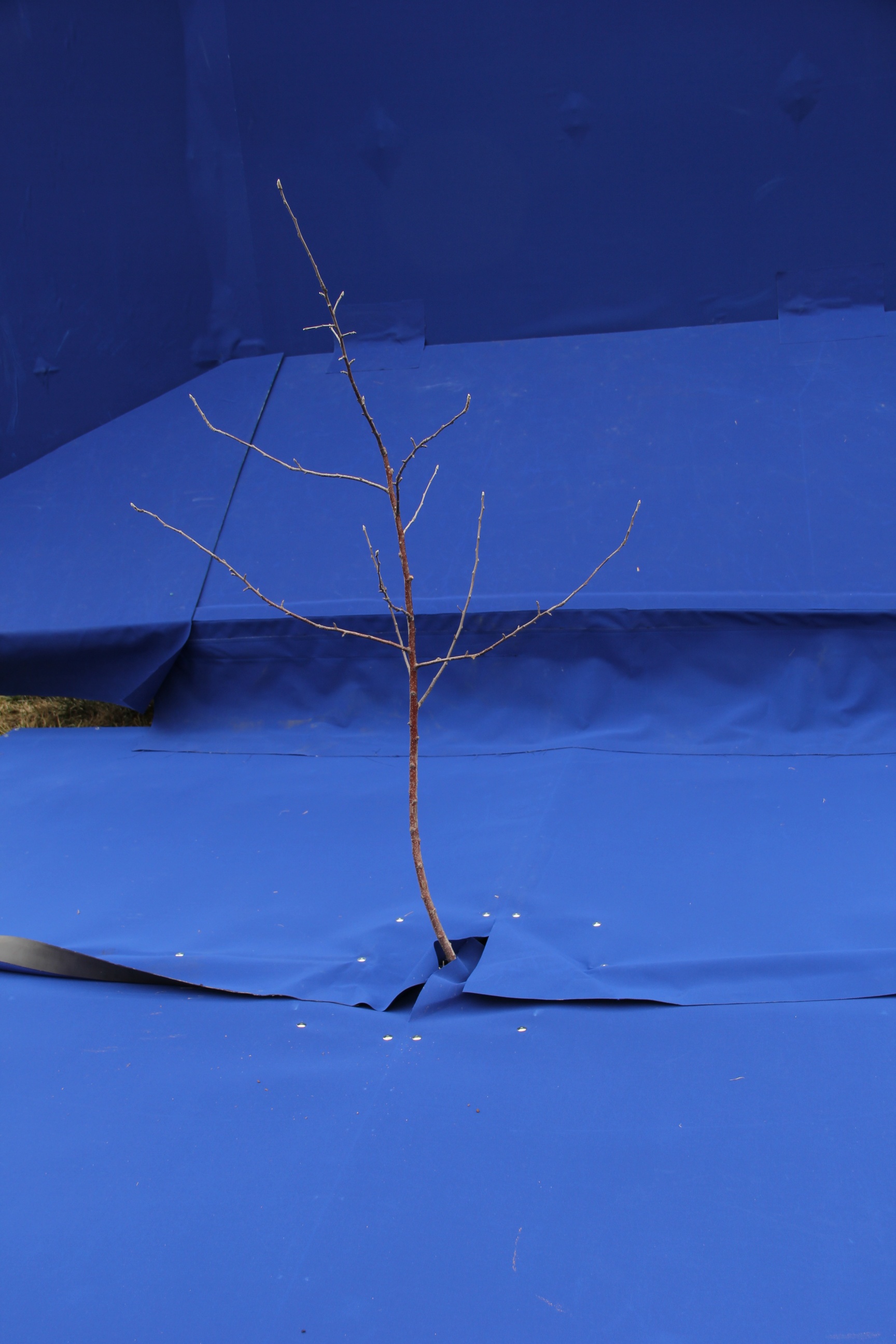}   } 
\end{minipage}
\begin{minipage}[b]{.24\linewidth}
\subfloat[Reconstruction]  
	{  
	\includegraphics[trim = 0.1cm 3cm 1cm 1cm,clip=true, height=0.28\textheight]{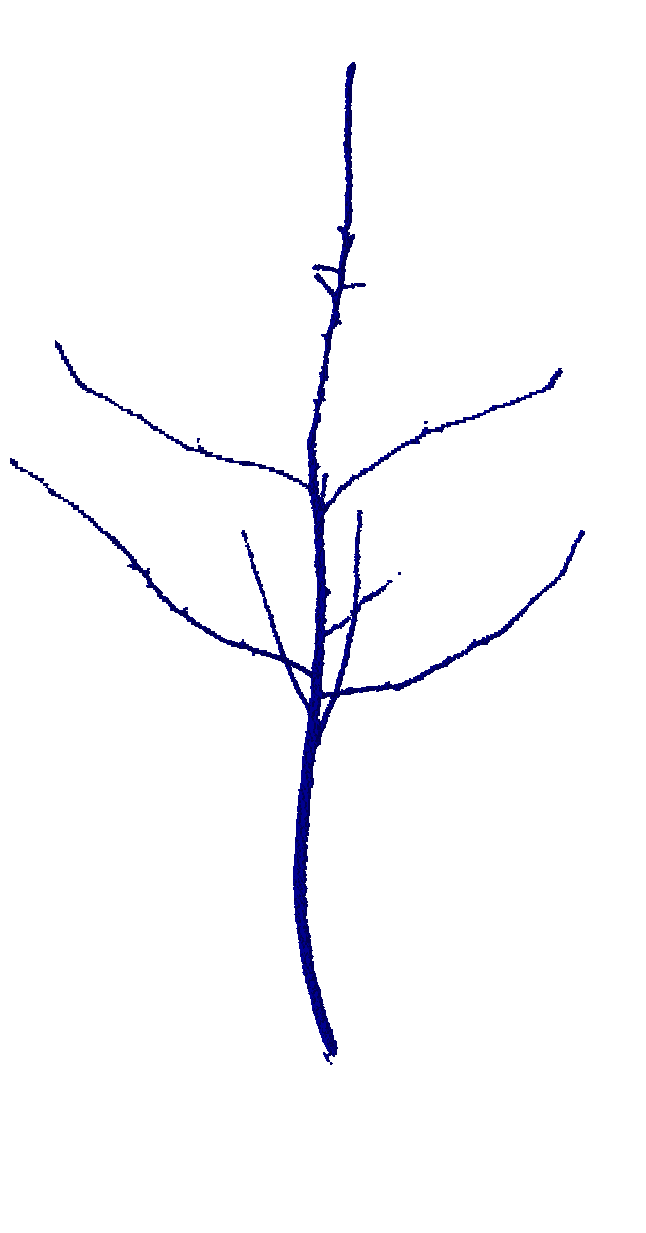}  

	}
\end{minipage}
\begin{minipage}[b]{.24\linewidth}
\subfloat[Curve skeleton]  
	{  
	\includegraphics[trim = 0.1cm 3cm 1cm 1cm,height=0.28\textheight]{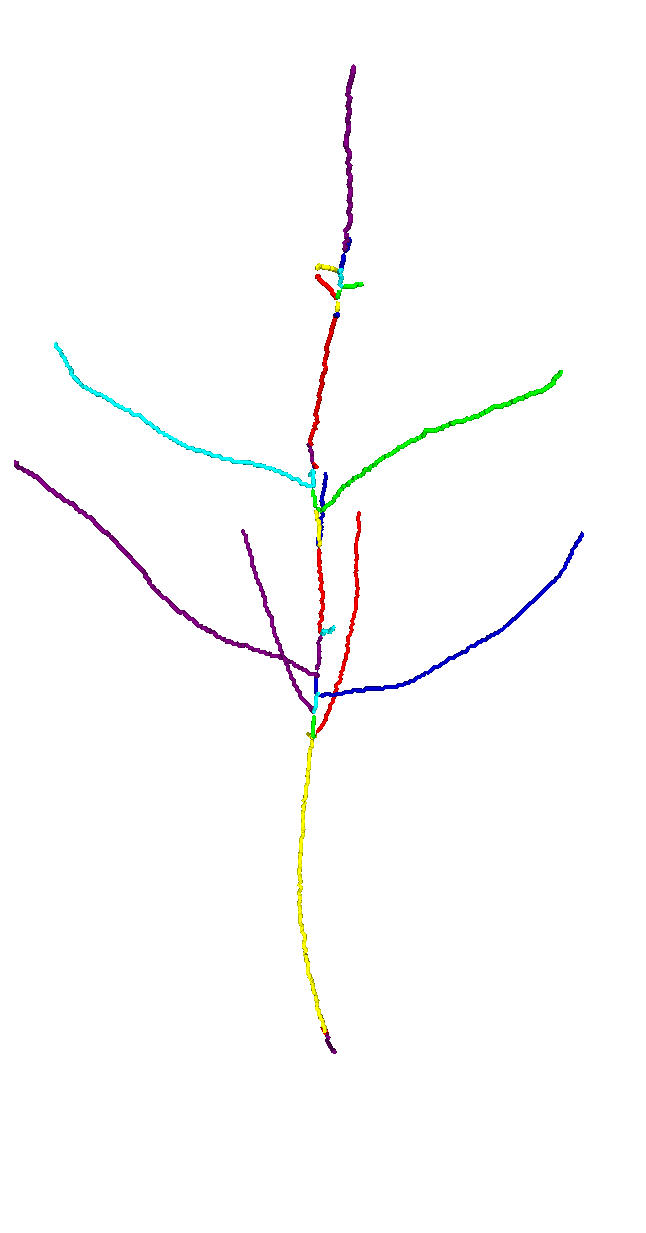}  

	}
\end{minipage}
\begin{minipage}[b]{.24\linewidth}
\subfloat[Graph representation]  
	{ 
	\includegraphics[trim = 0.1cm 3cm 1cm 1cm,height=0.28\textheight]{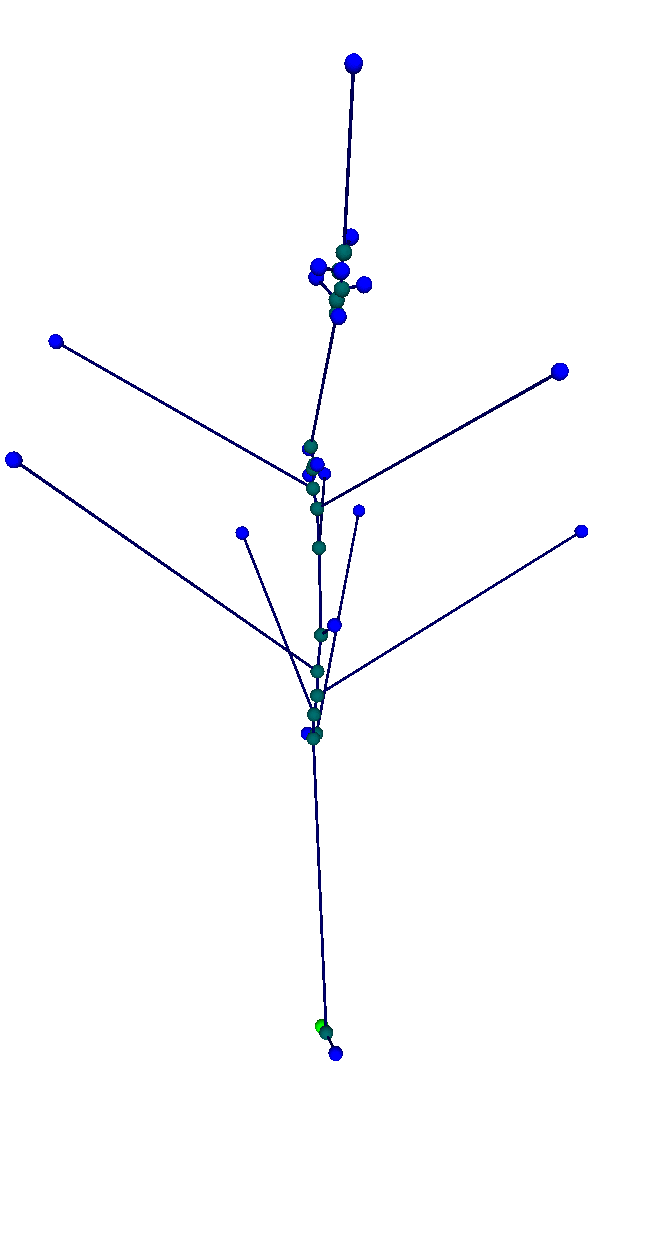}  
 
 }
\end{minipage}
\caption{\textbf{[Best viewed in color.]} Results for tree B.  No ground truth available. }
\label{fig:treeB_nogt}
\end{figure*}

\begin{figure*}[!ht]
\centering
\begin{minipage}[b]{.24\linewidth}
\subfloat[Full image of tree D]  
	    {  
	    \includegraphics[trim = 7.5cm 7.5cm 7.5cm 12.5cm, clip, height=0.28\textheight]{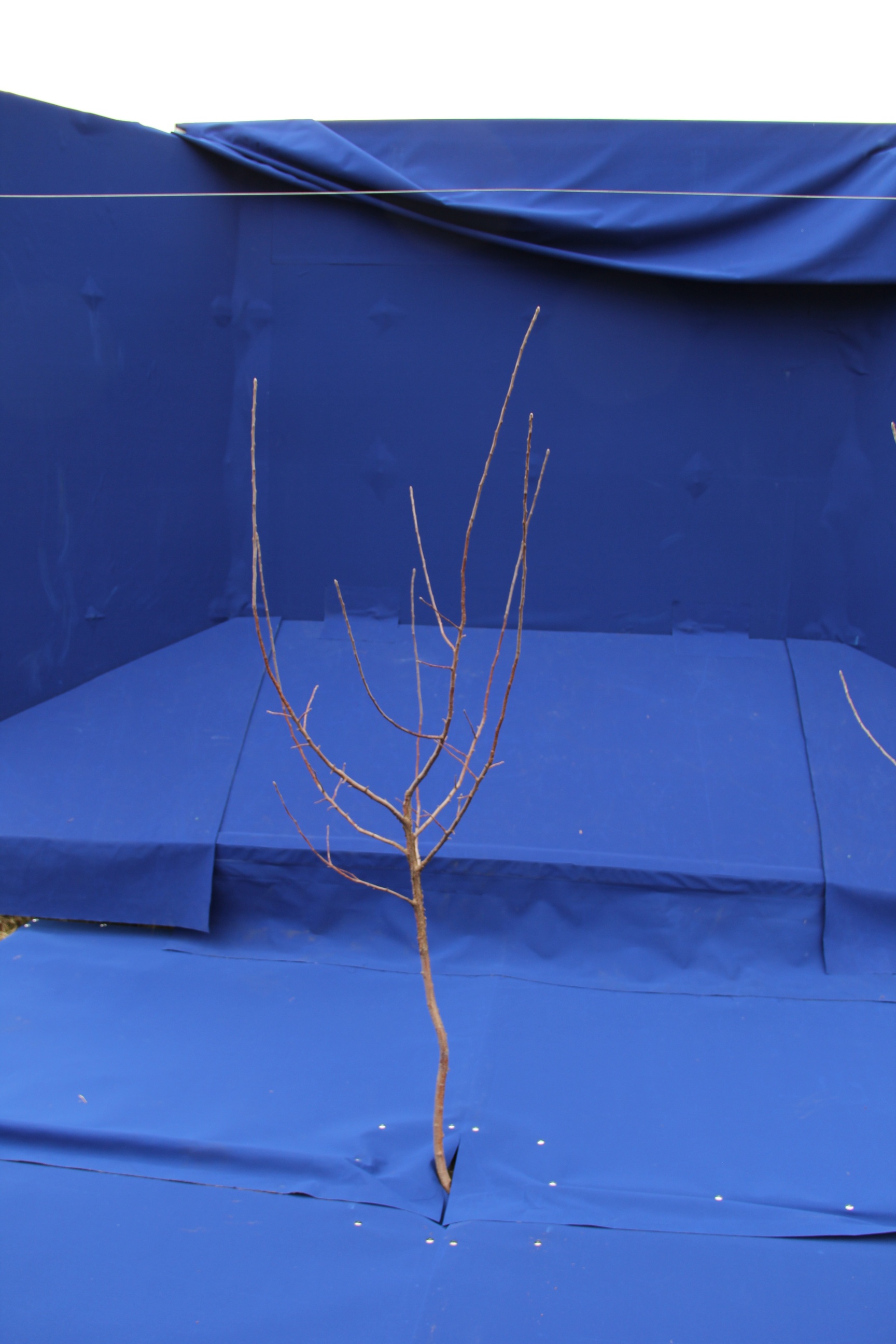}   } 
\end{minipage}
\begin{minipage}[b]{.24\linewidth}
\subfloat[Reconstruction]  
	{ 
	\includegraphics[trim = 3cm 3cm 3cm 1cm, clip, height=0.28\textheight]{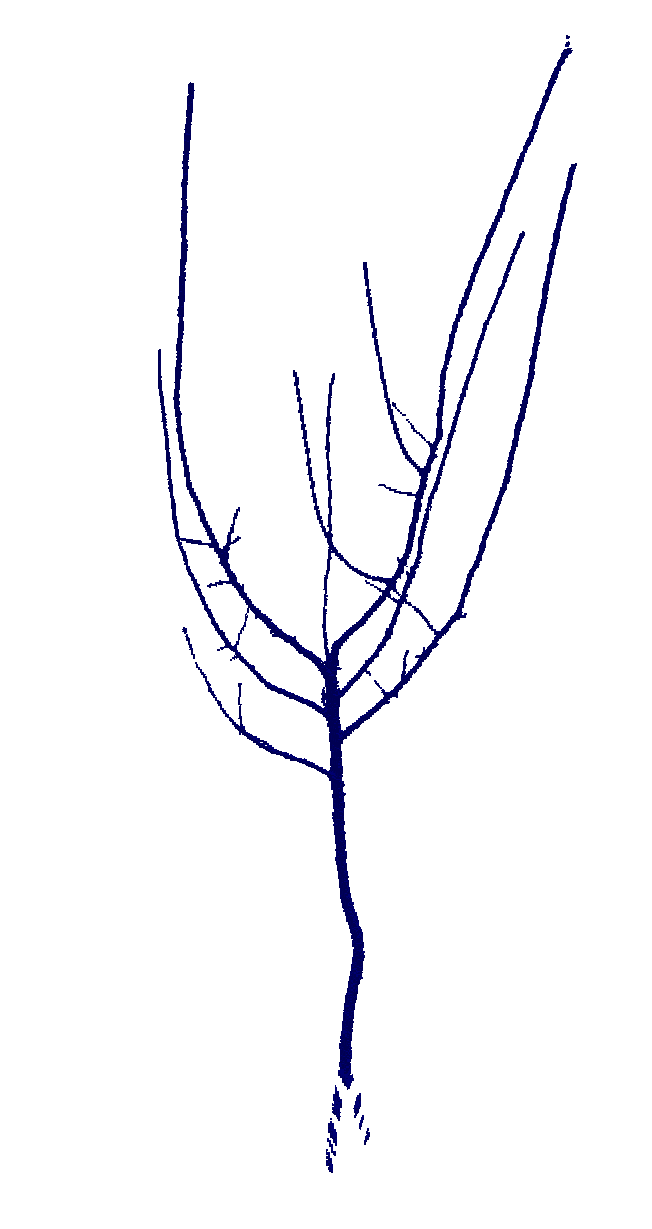}  

	}
\end{minipage}
\begin{minipage}[b]{.24\linewidth}
\subfloat[Curve skeleton]  
	{  
	\includegraphics[trim = 3cm 3cm 3cm 1cm, clip,height=0.28\textheight]{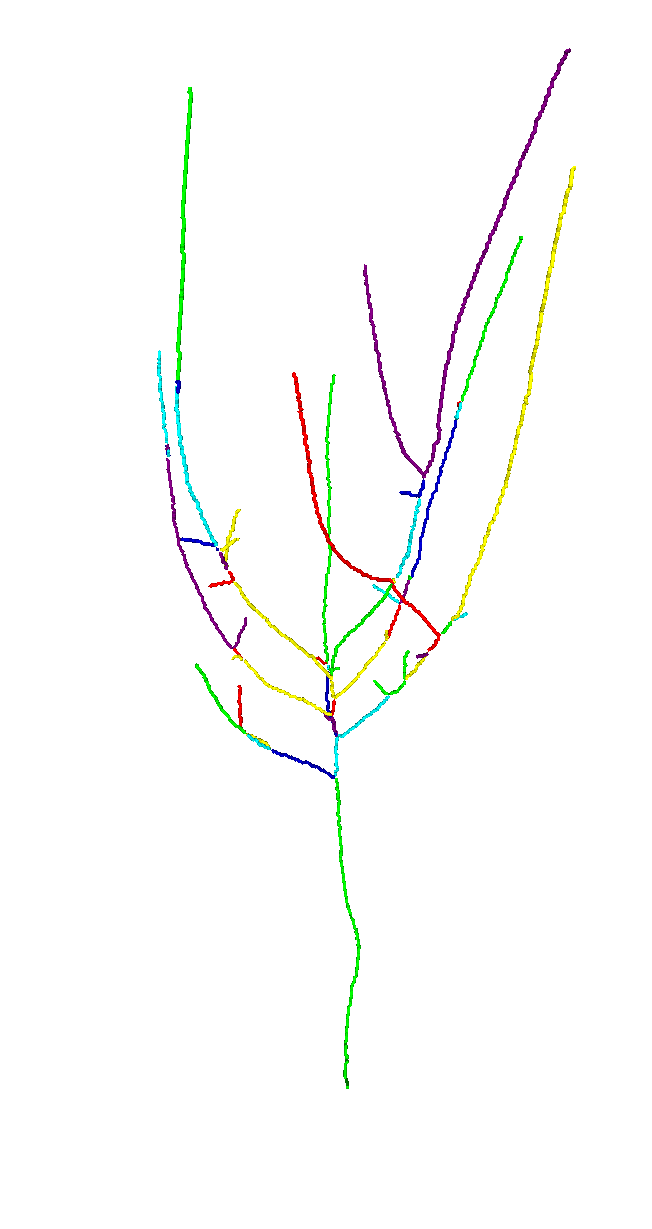}  

	}
\end{minipage}
\begin{minipage}[b]{.24\linewidth}
\subfloat[Graph representation]  
	{ 
	\includegraphics[trim = 3cm 3cm 3cm 1cm, clip,height=0.28\textheight]{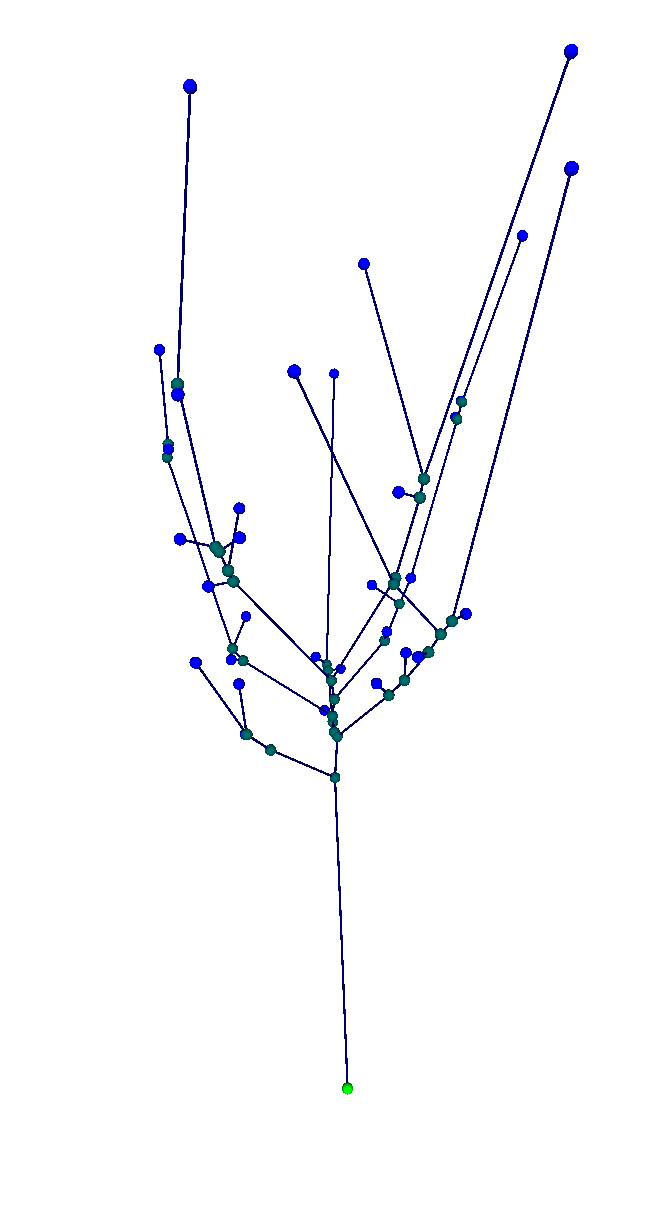}  
 
 }
\end{minipage}
\caption{\textbf{[Best viewed in color.]} Results for tree D.  No ground truth available.}
\label{fig:treeD_nogt}
\end{figure*}

\begin{figure*}[!ht]
\centering
\subfloat[Detail of tree D]  
	{  
	\includegraphics[trim = 3cm 3cm 3cm 1cm, clip, width=0.32\linewidth]{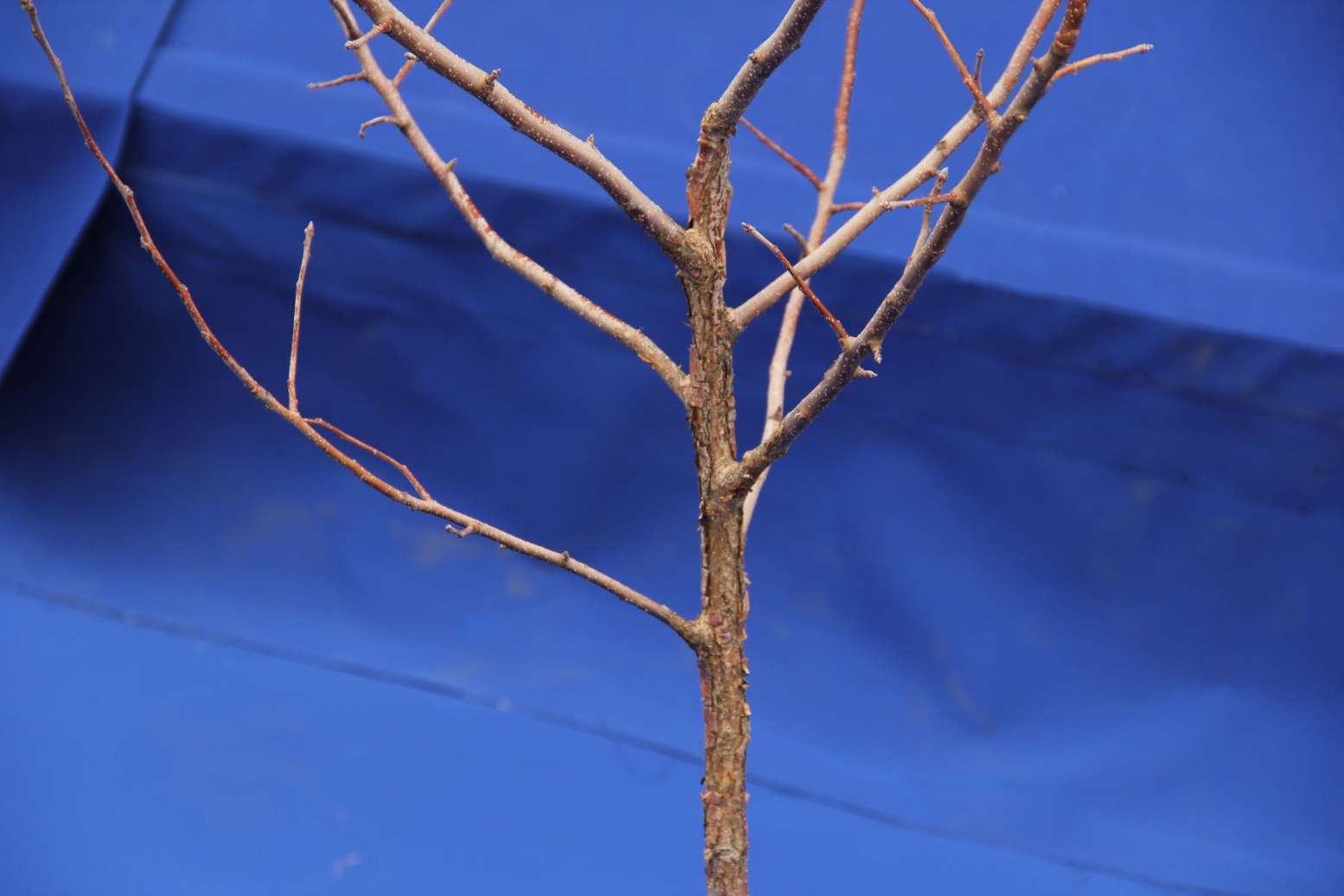}  

	}
\subfloat[Reconstruction]  
	{  
	\includegraphics[width=0.32\linewidth]{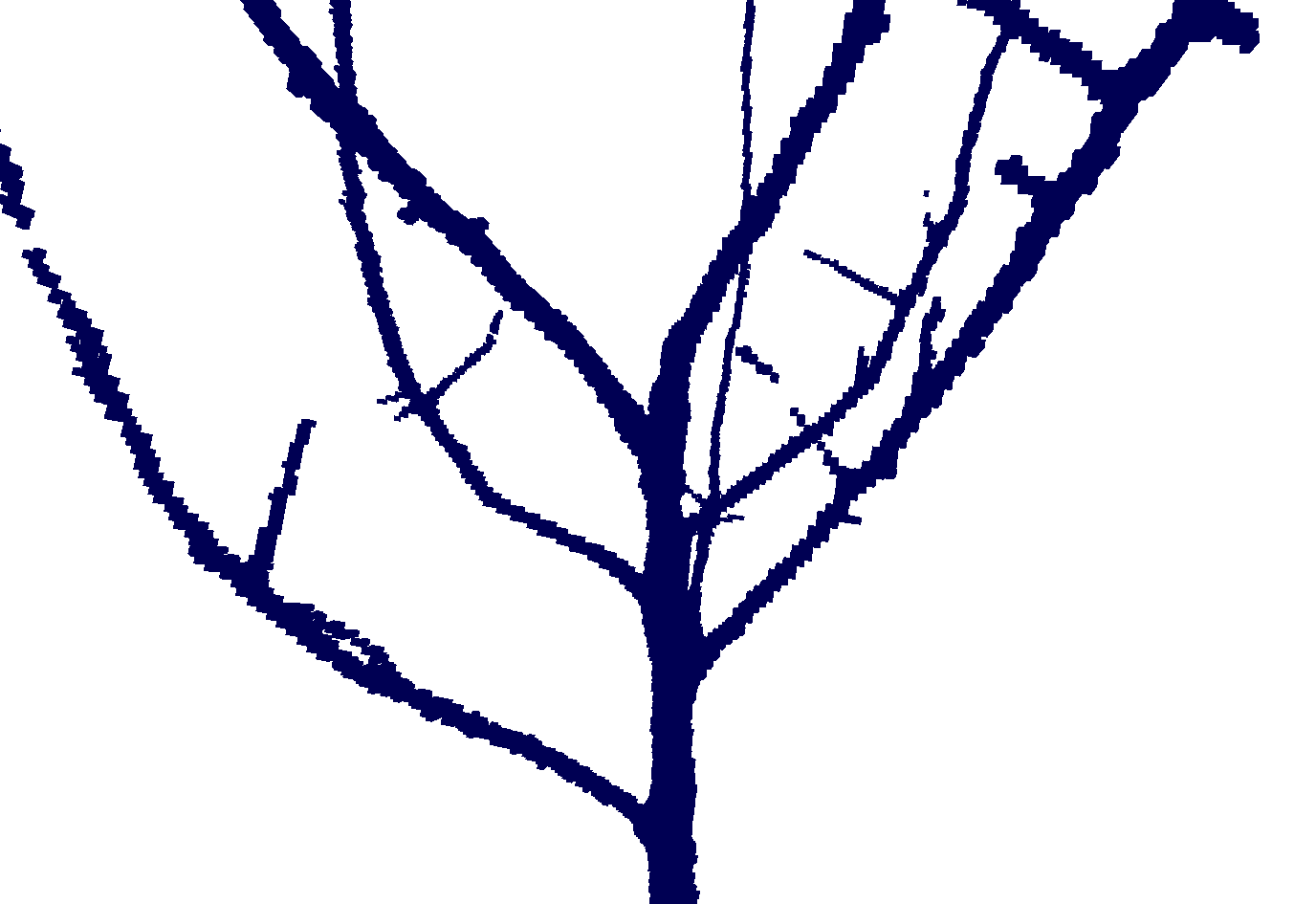}  

	}
\subfloat[Curve skeleton]  
	{ 
	\includegraphics[width=0.32\linewidth]{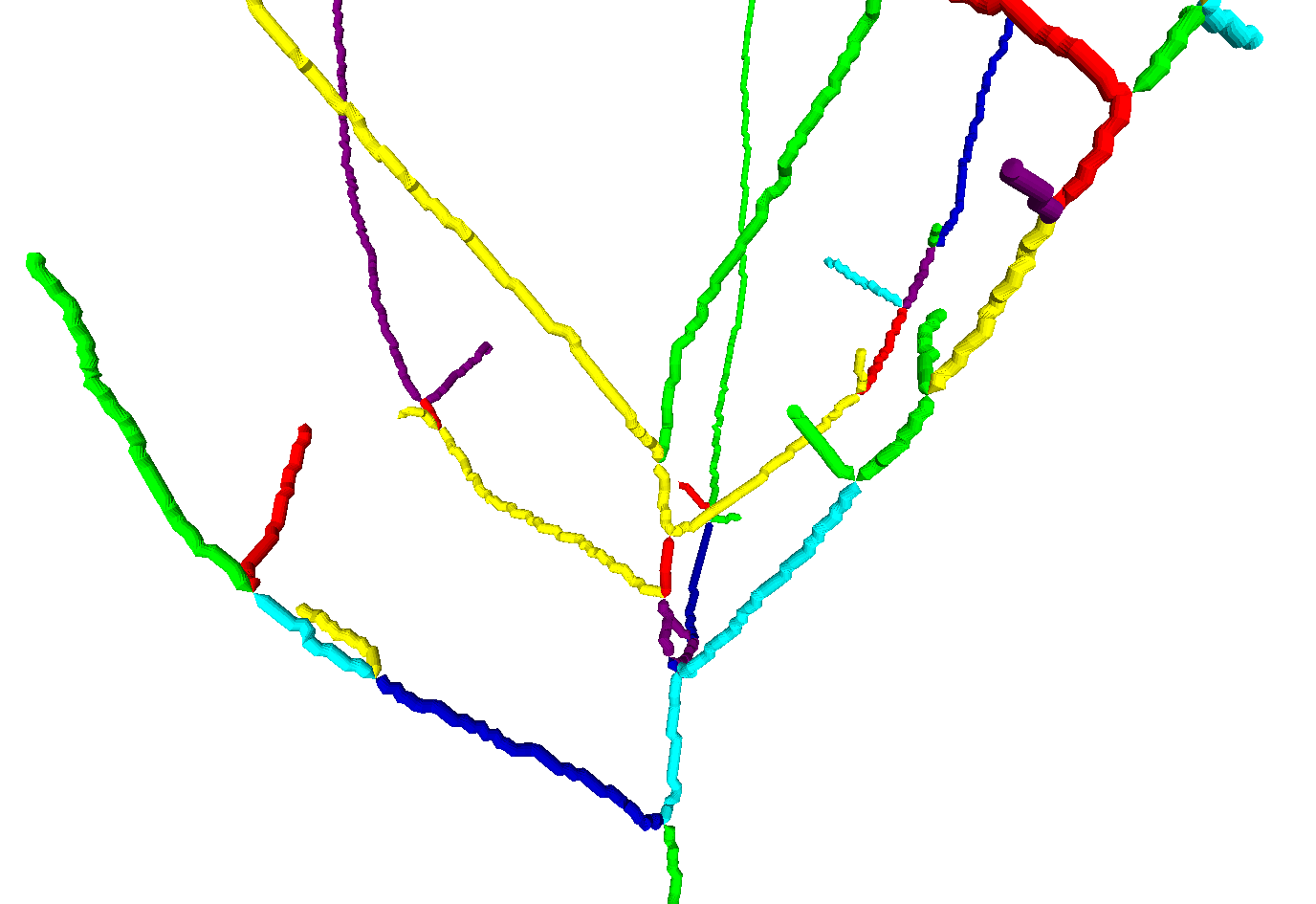}  \label{subfig:noisy_skeleton}
 }
\caption{\textbf{[Best viewed in color.]} A detail of the results for tree D.}
\label{fig:treeD_nogt_detail}
\end{figure*}

We qualitatively evaluate the robustness of our system using data from twelve trees in the orchard.  Figures \ref{fig:overview} and \ref{fig:treeB_nogt}-\ref{fig:treeD_nogt_detail} show some examples of the trees and their representation by RoTSE.  The three-dimensional nature of the trees can be seen in the video in our Supplementary Materials.\footnote{We intend to publish this video in an open access format at the National Agriculture Library's Ag Data Commons, for instance, so that it can serve as a companion to this paper.}  In general, the primary branches are captured by the system, while the secondary and higher-level branches are captured but less reliably.  A major reason for this is the voxel size setting; the high-order branches have smaller diameters, and in order to facilitate their capture the voxel size would have to be decreased, which would increase runtime. 

\subsection{Discussion}

In terms of accuracy, the methods involved in RoTSE utilize discrete divisions of the search region into voxels, and for the experiments in this paper, the minimum voxel size was set to $3$mm$^3$. Concerning the diameter measurements, the root mean squared error of $2.97$mm is almost the size of one of the voxel sides, which was a surprising result. The angle and length measurements were not as accurate, which may also be influenced by the discrete way in which those measurements were acquired.  For the angle measurements, the branch junction voxel as well as one voxel on the parent and child branches were used to compute the angle, but it may be that those particular voxels were not the most appropriate ones to use for computing the measurement.  On length, the individual voxel distances are summed to produce a branch length.  However, this may produce a length that is longer than the actual length because of the noisiness of the branch path (for example, notice the blue segment in the lower part of Figure \ref{subfig:noisy_skeleton}).  

On the subject of computational time, the main bottleneck for runtime is the reconstruction step.  The main parameters that affect runtime for the reconstruction method we use are the number of pixels, number of voxels, and the complexity of the object.  Since these are small trees, the initial voxel size for the reconstruction step has to be set relatively small at $12$mm$^3$. Small initial voxel sizes in conjunction with a relatively large search region because of varying distances from the truck to the tree, as well as differences in tree height and width, resulted in long run times for the reconstruction step.  

\section{Conclusions and Future Work}

We have described RoTSE, which creates shape models of leafless trees in field settings and computes tree traits such as branch junction locations, branch diameters, branch lengths, and branch angles.  The system was evaluated relative to accuracy in estimating tree traits, computation time, and robustness.  Through these evaluations, we could see that RoTSE at this point is suitable for structural phenotyping, and provides an advance in that field, considering that no in-field systems for structural phenotyping are currently available.  In the structural phenotyping problem, computational time is not as issue, but accuracy is.  The robotic pruning problem is an automation task that requires a fast runtime.  Consequently, the runtimes of RoTSE must be decreased in order for the system to be adequate for that purpose. 

In the course of this research, we have identified some future work that would ultimately improve RoTSE. First, the operation of the system as described in this paper is offline: images are acquired in the field for all trees, and then the images are transferred to a high performance computer, where the steps are run for all trees until completion. However, in our future work we intend to make the software steps more fully incorporated into the MRU, with the eventual goal of producing tree shape models and estimates in the field.  As a part of this work, we would like to speed up the reconstruction step, which is the major bottleneck in terms of runtime for the RoTSE system.  Finally, in this paper, we do not deal with cycles in the construction of the graph in Section \ref{sss:graph}.  However, in real-life situations, cycles are often present from branches overlapping each other and crossing, to intersections of the trellis materials with the tree.  Strategies to disambiguate branches segments that form cycles are part of our plans for future work.  

\subsubsection{Acknowledgements}
A. Tabb would like to acknowledge and thank Scott Wolford for his expertise and effort in the construction of the MRU and background units.  She would also like to thank Larry Crim for his expertise concerning the measurement of tree traits.

\bibliographystyle{IEEEtran}
\bibliography{tree_measurement_refs}

\begin{thebibliography}{10}
\providecommand{\url}[1]{#1}
\csname url@rmstyle\endcsname
\providecommand{\newblock}{\relax}
\providecommand{\bibinfo}[2]{#2}
\providecommand\BIBentrySTDinterwordspacing{\spaceskip=0pt\relax}
\providecommand\BIBentryALTinterwordstretchfactor{4}
\providecommand\BIBentryALTinterwordspacing{\spaceskip=\fontdimen2\font plus
\BIBentryALTinterwordstretchfactor\fontdimen3\font minus
  \fontdimen4\font\relax}
\providecommand\BIBforeignlanguage[2]{{%
\expandafter\ifx\csname l@#1\endcsname\relax
\typeout{** WARNING: IEEEtran.bst: No hyphenation pattern has been}%
\typeout{** loaded for the language `#1'. Using the pattern for}%
\typeout{** the default language instead.}%
\else
\language=\csname l@#1\endcsname
\fi
#2}}

\bibitem{Lehnert2015automated}
R.~Lehnert, ``Automated pruning with robotics,'' \emph{Good Fruit Grower},
  2015.

\bibitem{houle2010phenomics}
D.~Houle, D.~R. Govindaraju, and S.~Omholt, ``Phenomics: the next challenge,''
  \emph{Nature reviews genetics}, vol.~11, no.~12, pp. 855--866, 2010.

\bibitem{Li2014Review}
L.~Li, Q.~Zhang, and D.~Huang, ``A review of imaging techniques for plant
  phenotyping,'' \emph{Sensors}, vol.~14, no.~11, pp. 20\,078--20\,111, 2014.

\bibitem{sinoquet1997assessment}
H.~Sinoquet, P.~Rivet, and C.~Godin, ``Assessment of the three-dimensional
  architecture of walnut trees using digitising,'' \emph{Silva Fennica},
  vol.~31, no.~3, pp. 265--273, 1997.

\bibitem{arikapudi2015orchard}
R.~Arikapudi, S.~Vougioukas, and T.~Saracoglu, ``Orchard tree digitization for
  structural-geometrical modeling,'' in \emph{Precision agriculture'15}.\hskip
  1em plus 0.5em minus 0.4em\relax Wageningen Academic Publishers, 2015, pp.
  161--168.

\bibitem{vougioukas2016study}
S.~G. Vougioukas, R.~Arikapudi, and J.~Munic, ``A study of fruit reachability
  in orchard trees by linear-only motion,'' \emph{IFAC-PapersOnLine}, vol.~49,
  no.~16, pp. 277--280, 2016.

\bibitem{medeiros2016modeling}
\BIBentryALTinterwordspacing
H.~Medeiros, D.~Kim, J.~Sun, H.~Seshadri, S.~A. Akbar, N.~M. Elfiky, and
  J.~Park, ``Modeling dormant fruit trees for agricultural automation,''
  \emph{Journal of Field Robotics}, pp. n/a--n/a, 2016. [Online]. Available:
  \url{http://dx.doi.org/10.1002/rob.21679}
\BIBentrySTDinterwordspacing

\bibitem{livny2010automatic}
\BIBentryALTinterwordspacing
Y.~Livny, F.~Yan, M.~Olson, B.~Chen, H.~Zhang, and J.~El-Sana, ``Automatic
  reconstruction of tree skeletal structures from point clouds,'' \emph{ACM
  Trans. Graph.}, vol.~29, no.~6, pp. 151:1--151:8, Dec. 2010. [Online].
  Available: \url{http://doi.acm.org/10.1145/1882261.1866177}
\BIBentrySTDinterwordspacing

\bibitem{liu2012image}
W.~Liu, G.~Kantor, F.~D. la~Torre, and N.~Zheng, ``Image-based tree pruning,''
  in \emph{2012 IEEE International Conference on Robotics and Biomimetics
  (ROBIO)}, Dec 2012, pp. 2072--2077.

\bibitem{karkee2014identification}
M.~Karkee, B.~Adhikari, S.~Amatya, and Q.~Zhang, ``Identification of pruning
  branches in tall spindle apple trees for automated pruning,'' \emph{Computers
  and Electronics in Agriculture}, vol. 103, pp. 127--135, 2014.

\bibitem{karkee2015method}
M.~Karkee and B.~Adhikari, ``A method for three-dimensional reconstruction of
  apple trees for automated pruning,'' \emph{Transactions of the ASABE},
  vol.~58, no.~3, pp. 565--574, 2015.

\bibitem{elfiky2015automation}
N.~M. Elfiky, S.~A. Akbar, J.~Sun, J.~Park, and A.~Kak, ``Automation of dormant
  pruning in specialty crop production: An adaptive framework for automatic
  reconstruction and modeling of apple trees,'' in \emph{Proceedings of the
  IEEE Conference on Computer Vision and Pattern Recognition Workshops}, 2015,
  pp. 65--73.

\bibitem{akbar2016novel}
S.~A. Akbar, N.~M. Elfiky, and A.~Kak, ``A novel framework for modeling dormant
  apple trees using single depth image for robotic pruning application,'' in
  \emph{Robotics and Automation (ICRA), 2016 IEEE International Conference
  on}.\hskip 1em plus 0.5em minus 0.4em\relax IEEE, 2016, pp. 5136--5142.

\bibitem{Chattopadhyay2016Measuring}
S.~Chattopadhyay, S.~A. Akbar, N.~M. Elfiky, H.~Medeiros, and A.~Kak,
  ``Measuring and modeling apple trees using time-of-flight data for automation
  of dormant pruning applications,'' in \emph{2016 IEEE Winter Conference on
  Applications of Computer Vision (WACV)}, March 2016, pp. 1--9.

\bibitem{botterill2016robot}
\BIBentryALTinterwordspacing
T.~Botterill, S.~Paulin, R.~Green, S.~Williams, J.~Lin, V.~Saxton, S.~Mills,
  X.~Chen, and S.~Corbett-Davies, ``A robot system for pruning grape vines,''
  \emph{Journal of Field Robotics}, pp. n/a--n/a, 2016. [Online]. Available:
  \url{http://dx.doi.org/10.1002/rob.21680}
\BIBentrySTDinterwordspacing

\bibitem{meshlab}
P.~Cignoni, M.~Callieri, M.~Corsini, M.~Dellepiane, F.~Ganovelli, and
  G.~Ranzuglia, ``{MeshLab: an Open-Source Mesh Processing Tool},'' in
  \emph{Eurographics Italian Chapter Conference}, V.~Scarano, R.~D. Chiara, and
  U.~Erra, Eds.\hskip 1em plus 0.5em minus 0.4em\relax The Eurographics
  Association, 2008.

\bibitem{tabb2017solving}
\BIBentryALTinterwordspacing
A.~Tabb and K.~M. Ahmad Yousef, ``Solving the robot-world hand-eye(s)
  calibration problem with iterative methods,'' \emph{Machine Vision and
  Applications}, May 2017. [Online]. Available:
  \url{http://dx.doi.org/10.1007/s00138-017-0841-7}
\BIBentrySTDinterwordspacing

\bibitem{tabb2017solving_dataset}
A.~Tabb. (2017) Data from: Solving the robot-world hand-eye(s) calibration
  problem with iterative methods.
  \url{http://dx.doi.org/10.15482/USDA.ADC/1340592}.

\bibitem{tabb2017automatic}
A.~Tabb and H.~Medeiros, ``Automatic segmentation in dynamic outdoor
  environments,'' \emph{arXiv:1702.07611 [cs.CV]}, 2017.

\bibitem{tabb2013shape}
A.~Tabb, ``Shape from silhouette probability maps: reconstruction of thin
  objects in the presence of silhouette extraction and calibration error,'' in
  \emph{Computer Vision and Pattern Recognition (CVPR), 2013 IEEE Conference
  on}, June 2013.

\bibitem{tabb2014shape}
------, ``Shape from inconsistent silhouette: Reconstruction of objects in the
  presence of segmentation and camera calibration error,'' Ph.D. dissertation,
  Purdue University, 2014.

\bibitem{tabb2017fast}
A.~Tabb and H.~Medeiros, ``Fast and robust curve skeletonization for real-world
  elongated objects,'' \emph{arXiv:1702.07619 [cs.CV]}, 2017.

\bibitem{meijster2002general}
A.~Meijster, J.~B. Roerdink, and W.~H. Hesselink, ``A general algorithm for
  computing distance transforms in linear time,'' in \emph{Mathematical
  Morphology and its applications to image and signal processing}.\hskip 1em
  plus 0.5em minus 0.4em\relax Springer, 2002, pp. 331--340.

\end{thebibliography}

\section{Erratum}\label{erratum}

We regret that there were some errors in the original paper, concerned with the computation of the accuracy of RoTSE in Table \ref{table:accuracy}. The original table was \ref{table:original}:

\begin{table}[!h]
\caption{Summary of the accuracy of RoTSE for 2 trees. MSE is the mean squared error and STD is the standard deviation of the error.}
\begin{center}
{
\begin{tabular}{l||c|c|c}
\hline 
 & Diameter (mm) & Length (mm) & Angle (degrees)\tabularnewline
\hline 
\hline 
MSE & 0.99 & 45.64 & 10.36\tabularnewline
\hline 
STD & 2.85 & 124.51 & 32.18\tabularnewline
\hline 
\end{tabular}
}
\end{center}
\label{table:original}
\end{table} 

and it has been corrected to Table \ref{table:corrected}:

\begin{table}[!h]
\caption{Summary of the accuracy of RoTSE for 2 trees. RMSE is the root mean squared error and STD is the standard deviation of the error.}
\begin{center}
{
\begin{tabular}{l||c|c|c}
\hline 
 & Diameter (mm) & Length (mm) & Angle (degrees)\tabularnewline
\hline 
\hline 
RMSE & 2.97 & 136.92 & 31.07\tabularnewline
\hline 
STD & 2.85 & 124.51 & 32.18\tabularnewline
\hline 
\end{tabular}
}
\end{center}
\label{table:corrected}
\end{table} 

We have also correct the text when these numbers appear.

\end{document}